\documentclass[10pt,twocolumn,letterpaper]{article}

\usepackage[pagenumbers]{cvpr} 

\usepackage{bbding}
\usepackage[misc]{ifsym}
\usepackage{multirow}

\definecolor{cvprblue}{rgb}{0.21,0.49,0.74}
\usepackage[pagebackref,breaklinks,colorlinks,allcolors=cvprblue]{hyperref}
\usepackage{array}
\usepackage{xcolor}

\definecolor{darkgreen}{rgb}{0.0, 0.5, 0.0} 
\definecolor{mambaPurple}{rgb}{197, 153, 204}
\usepackage{CJK}
\usepackage{enumitem}
\usepackage{pifont}
\usepackage{adjustbox}

\definecolor{hatsuneGreen}{RGB}{0, 163, 175}

\makeatletter
\def\blfootnote{\xdef\@thefnmark{}\@footnotetext}
\makeatother

\title{\raisebox{-2mm}{\includegraphics[width=0.12\textwidth]{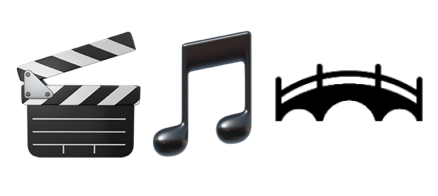}}%
    \hspace{1mm}Multimodal Music Generation with Explicit Bridges and \\ Retrieval Augmentation
}

\author{
    Baisen Wang$^{1,2}$ \qquad
    Le Zhuo$^3$ \qquad
    Zhaokai Wang$^{4,3}$ \qquad
    Chenxi Bao$^5$ \qquad
    Chengjing Wu$^6$ \qquad
    \\
    Xuecheng Nie$^6$ \qquad
    Jiao Dai$^{1,2}$ \qquad 
    Jizhong Han$^{1,2}$ \qquad 
    Yue Liao$^{7\dag}$ \qquad
    Si Liu$^{8\dag}$
    \\
    \fontsize{11pt}{12pt}\selectfont
    $^1$Institute of Information Engineering, Chinese Academy of Sciences\\
    \fontsize{11pt}{12pt}\selectfont
    $^2$School of Cyberspace Security, University of Chinese Academy of Sciences\\
    \fontsize{11pt}{12pt}\selectfont
    $^3$Shanghai AI Laboratory\qquad
    $^4$Shanghai Jiao Tong University 
    \qquad
    \fontsize{11pt}{12pt}\selectfont
    $^5$University of Edinburgh\\
    \fontsize{11pt}{12pt}\selectfont
    $^6$MT Lab, Meitu Inc.\qquad
    $^7$The Chinese University of Hong Kong\qquad
    $^8$Beihang University\qquad
    \\
    {\tt\small \{wbs2788, zhuole1025, zhaokaiwang99, liaoyue.ai\}@gmail.com
    }
}

\begin{document}
\maketitle
\blfootnote{$^{\dag}$Corresponding authors.}

\begin{abstract}

Multimodal music generation aims to produce music from diverse input modalities, including text, videos, and images. Existing methods use a common embedding space for multimodal fusion. Despite their effectiveness in other modalities, their application in multimodal music generation faces challenges of data scarcity, weak cross-modal alignment, and limited controllability. This paper addresses these issues by using explicit bridges of text and music for multimodal alignment.
We introduce a novel method named Visuals Music Bridge (VMB). Specifically, a Multimodal Music Description Model converts visual inputs into detailed textual descriptions to provide the text bridge; a Dual-track Music Retrieval module that combines broad and targeted retrieval strategies to provide the music bridge and enable user control. Finally, we design an Explicitly Conditioned Music Generation framework to generate music based on the two bridges. We conduct experiments on video-to-music, image-to-music, text-to-music, and controllable music generation tasks, along with experiments on controllability. The results demonstrate that VMB significantly enhances music quality, modality, and customization alignment compared to previous methods. VMB sets a new standard for interpretable and expressive multimodal music generation with applications in various multimedia fields. Demos and code are available at \url{https://github.com/wbs2788/VMB}.
\end{abstract}    
\section{Introduction}
\label{sec:intro}

Music transcends linguistic and cultural barriers, uniquely resonating with human emotions and fostering connections beyond words.
Typically, when people experience visual, textual, or auditory inputs in their daily lives, they instinctively associate them with certain music motifs. This inherent ability to link sensory experiences with music underscores the potential for a system designed to generate music from various modalities.

\begin{figure*}[t]
  \centering
   \includegraphics[width=0.9\linewidth]{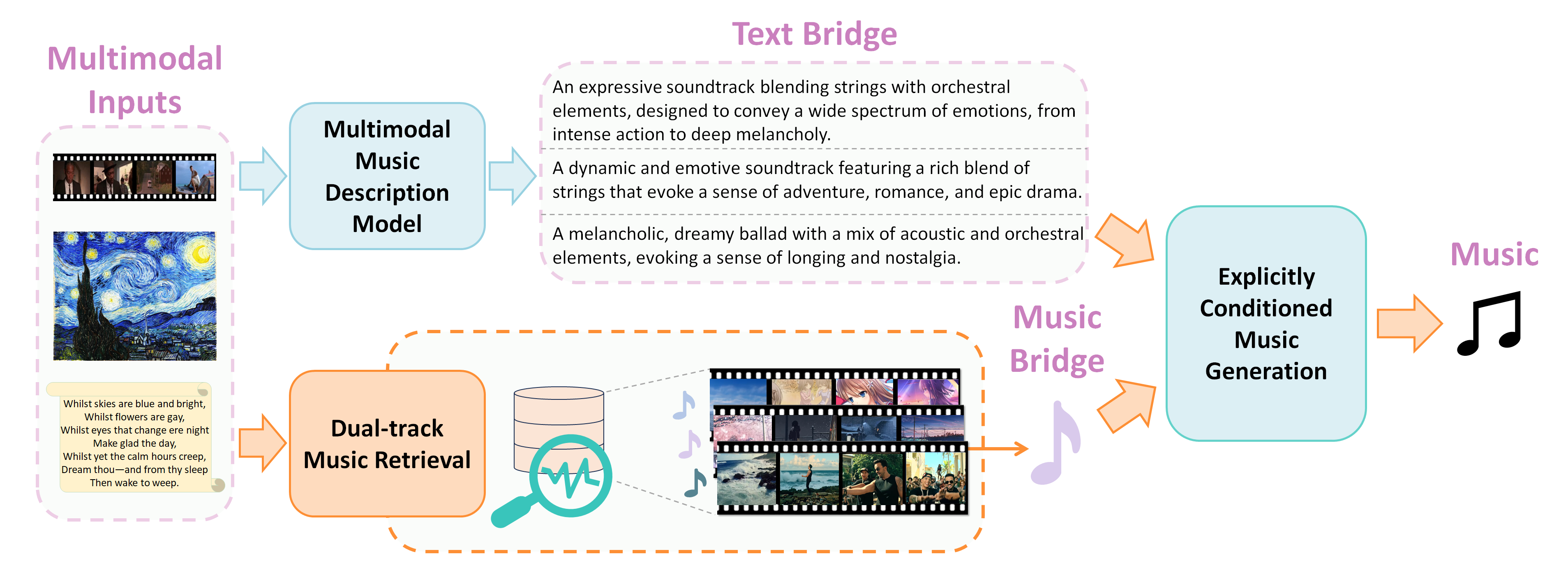}
    \vspace{-3mm}
   \caption{\textbf{Overview of the VMB framework.} We employ text and music as two explicit bridges for multimodal music generation. Text-form music description is obtained with the Multimodal Music Description model. Reference music is retrieved with the Dual-track Music Retrieval module. The two bridges are fed into the Explicitly Conditioned Music Generation module to generate output music. }\vspace{-3mm}
   \label{fig:overview}
\end{figure*}

Recent advancements in generative models have led to significant progress in producing music mainly from textual descriptions,~\emph{i.e.}, text-to-music generation~\citep{musiclm, musicgen, stableaudio} that translate the semantics conveyed in text into corresponding music. However, extending music generation to other modalities, such as images~\cite{wang2023continuous, chowdhury2024melfusion} or videos~\citep{cmt, musprod, video2music}, is still challenging and remain in its preliminary stage. For the more challenging task \emph{multimodal music generation} that takes multiple input modalities (\emph{e.g.}, text, image, video) to generate music, existing methods mostly use a common embedding space as the bridge between multiple modalities~\cite{m2ugen, VLM:NeXTGPT, codi}. Such approaches often struggle with maintaining high quality and accurate modality correlations in their musical outputs.  

We argue that several key limitations hinder the development of effective multimodal music generation systems. Firstly, 
for modalities other than text, the scarcity of large-scale, high-quality music-paired datasets hinders the learning of meaningful cross-modal relationships. 
Secondly, different modalities contribute to music generation in distinct and complementary ways,~\emph{e.g.}, text offers explicit semantic cues such as themes and emotions, while images and videos encapsulate visual emotions, atmosphere, style, and fine-grained temporal dynamics like rhythm.
This highlights the importance of stronger multimodal alignment. 
Thirdly, existing methods using common embedding space as the intermedia often lack interpretability and fine-grained controllability, limiting the ability to manipulate specific musical attributes such as instruments, style, or rhythm.

To address these challenges, we propose to \textbf{utilize text and music as explicit bridges} for multimodal alignment. In contrast to previous methods that use a joint embedding space for implicit alignment, the explicit bridge helps to mitigate the data scarcity issue and improve multimodal alignment by utilizing abundant text-music paired data. For text-bridge, we convert images and videos into detailed music descriptions with a multimodal music description model that guides the music generation process. For music-bridge, we introduce retrieval-augmented generation (RAG) into multimodal music synthesis to condition the model on relevant music pieces with both broad thematic alignment and targeted attribute control. Text and music complement each other with their distinct roles in multimodal alignment. This method also enhances controllability as users can specify the output music by modifying the text description, or even providing their own music for reference.
    
Based on the above design principles, we design a novel multimodal music generation framework, namely Visuals Music Bridge (VMB), that supports creating high-quality music from various modalities including text, audio, video, and music. As illustrated in Fig.~\ref{fig:overview}, our VMB framework consists of the following core components:

\begin{enumerate}
    \item \textbf{Multimodal Music Description Model}: We curate a multimodal music dataset consisting of video-music-description triplets and various musical attribute annotations,~\emph{e.g.}, genre, mood, and instruments. Given this dataset, we introduce the Multimodal Music Description Model, which is built upon InternVL2~\cite{VLM:InternVL-1.5} to accurately interpret visual inputs and translate them into detailed musical descriptions in natural language, serving as the text bridge for music generation. 

    \item \textbf{Dual-track Music Retrieval}: We design a Dual-track Music Retrieval framework tailored for multimodal music generation, which retrieves relevant music pieces leveraging both \emph{broad} and \emph{targeted} retrieval strategies. Broad retrieval identifies overall alignments of emotional and thematic content, establishing global coherence, while targeted retrieval focuses on specific musical attributes like tempo, instrumentation, and genre, allowing users to customize individual elements of the music.

    \item \textbf{Explicitly Conditioned Music Generation}: Given the text bridge and the music bridge, we propose the architecture of Explicitly Conditioned Music Generation, which integrates the two explicit bridges into a text-to-music diffusion transformer. Our model employs Music ControlFormer to integrate fine-grained control from the broad retrieval and leverages Stylization Module for overall conditions from the targeted retrieval.
\end{enumerate}

We conduct extensive experiments on video-to-music, text-to-music, image-to-music, and controllable music generation tasks. The results demonstrate that VMB is a robust, flexible framework that improves music quality, enhances alignment between input modalities and generated music, and offers high controllability. Our approach represents a step toward sophisticated and accessible multimodal music generation, with broad applications in multimedia.
\section{Related Work}

\begin{figure*}[t]
  \centering
   \includegraphics[width=\linewidth]{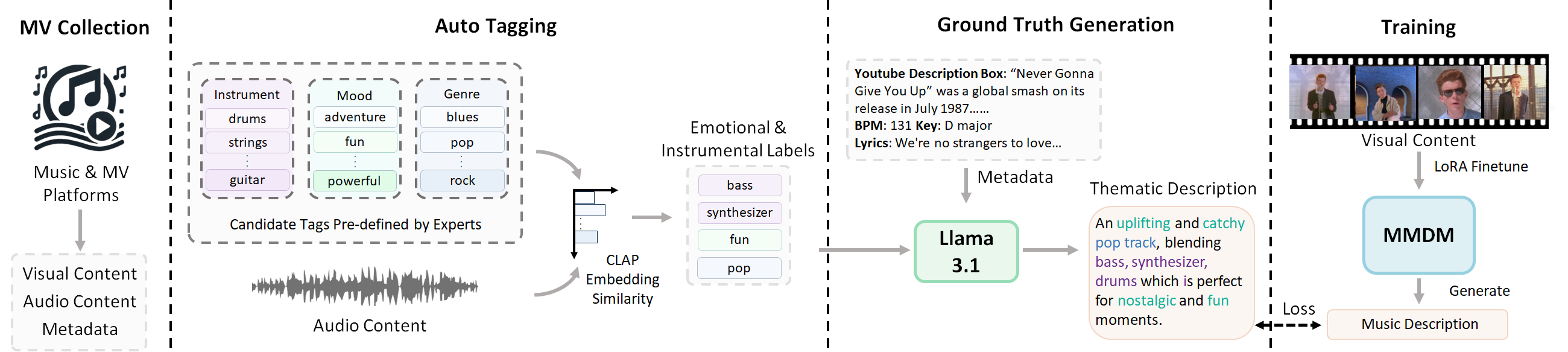}

   \caption{\textbf{Pipeline of the Multimodal Music Description Model (MMDM).} This process starts with the collection of music videos, followed by automated tagging to refine audio annotations using CLAP embedding similarities. Metadata and thematic descriptions are synthesized by the Llama 3.1 model to create training targets. The training utilizes LoRA fine-tuning in the MMDM to transform multimodal inputs into targeted music descriptions that align with the visual content's themes.}
   \label{fig:3.1}
\end{figure*}

\noindent\textbf{Multi-modal Music Generation.} 
In recent years, automatic music generation has witnessed significant advancements in symbolic form~\cite{musegan, midinet, musicvae, music_transformer, remi} and audio form~\cite{stableaudio, audioldm, noise2music}.
Meanwhile, many efforts has been made towards conditional music generation from other input modalities. Text-to-music methods aim to generate music from text descriptions and have made notable progress, such as AudioLDM~\cite{audioldm}, MusicLM~\cite{musiclm}, MusicGen~\cite{musicgen}, Stable Audio Open~\cite{stableaudio}. 
Building on this, MeLFusion~\cite{chowdhury2024melfusion} enhances the generation process by using an image as a condition to assist the textual description in producing music.

For video-to-music generation, CMT~\cite{cmt}, an early attempt to generate background music from videos, uses rule-based rhythmic features to translate video dynamics into musical rhythms to address the challenge of sparse data. 
MIDI-based video soundtrack generation methods~\cite{cmt, musprod, video2music, diffbgm} leverage the characteristics of MIDI to extract rhythmic elements from videos, enabling the generation of music that aligns with the video's rhythm. Additionally, these methods incorporate music theory knowledge to assist in generating background music for videos.
Audio-based methods~\cite{v2meow, vidmuse}, unlike MIDI-based approaches that are constrained by the limited availability of MIDI data, take advantage of large-scale audio datasets. By leveraging a variety of pre-trained models, audio-based methods aim to uncover the complex connections between video content and music, addressing the challenges posed by the scarcity of large, labeled datasets.
However, video-music alignment in current methods is still unsatisfying due to the inherent intricate alignment between visual inputs and music outputs.

In contrast to these methods that generate music from a single input, multimodal music generation consider multiple input modalities. Some multimodal generation works like NExT-GPT~\cite{VLM:NeXTGPT} and CoDi~\cite{codi} does not treat music as a separate modality but as part of audio generation. Consequently, their composed music has limited quality. M$^2$UGen~\cite{m2ugen} proposes a multimodal music generation method for video-to-music, image-music and text-music tasks. 
However, the feature-based modality fusion approach, adopted by most multimodal music generation methods, fails to fully capture the complex visual-music relationships. Moreover, their controllability is limited as extracted features cannot be manipulated directly. In this work, we employ text and music as the explicit bridges for multimodal fusion, which proves to be a effective approach in multimodal music generation.

\noindent\textbf{Retrieval-Augmented Generation.} 
Retrieval-augmented generation (RAG) has become a common technique for enhancing the fidelity and diversity of language models with an additional knowledge base in the field of natural language processing~\cite{lewis2020retrieval,guu2020retrieval,zhang2023retrieve,caffagni2024wiki}. Due to its effectiveness, RAG is further generalized to diverse fields and tasks, such as visual recognition~\cite{long2022retrieval,hu2023reveal}, image generation~\cite{sheynin2022knn,chen2022re}, 3D generation~\cite{wu2022learning,wang2024themestation}, and drug discovery~\cite{aguilar2022fragment}. However, RAG in music generation has not yet been explored. To the best of our knowledge, we propose the first retrieval-augmented pipeline for multimodal music generation, which dynamically incorporates dual-form of musical knowledge into a pre-trained music generation model to bridge the modality gap, boost cross-modal generation performance, and increase controllability. 

\section{Methodology}

In this section, we present our multimodal music generation framework, consisting of a Multimodal Music Description Model that generates textual descriptions for emotional
and thematic framing, providing the text bridge; a Dual-track Music Retrieval module to retrieve relevant music, providing the music bridge; and an Explicitly Conditioned Music Generation module that leverages the two bridges to generate the final music.

\subsection{Multimodal Music Description Model}
\label{sec:mmcm}

\begin{figure*}[t]
  \centering
   \includegraphics[width=0.9\linewidth]{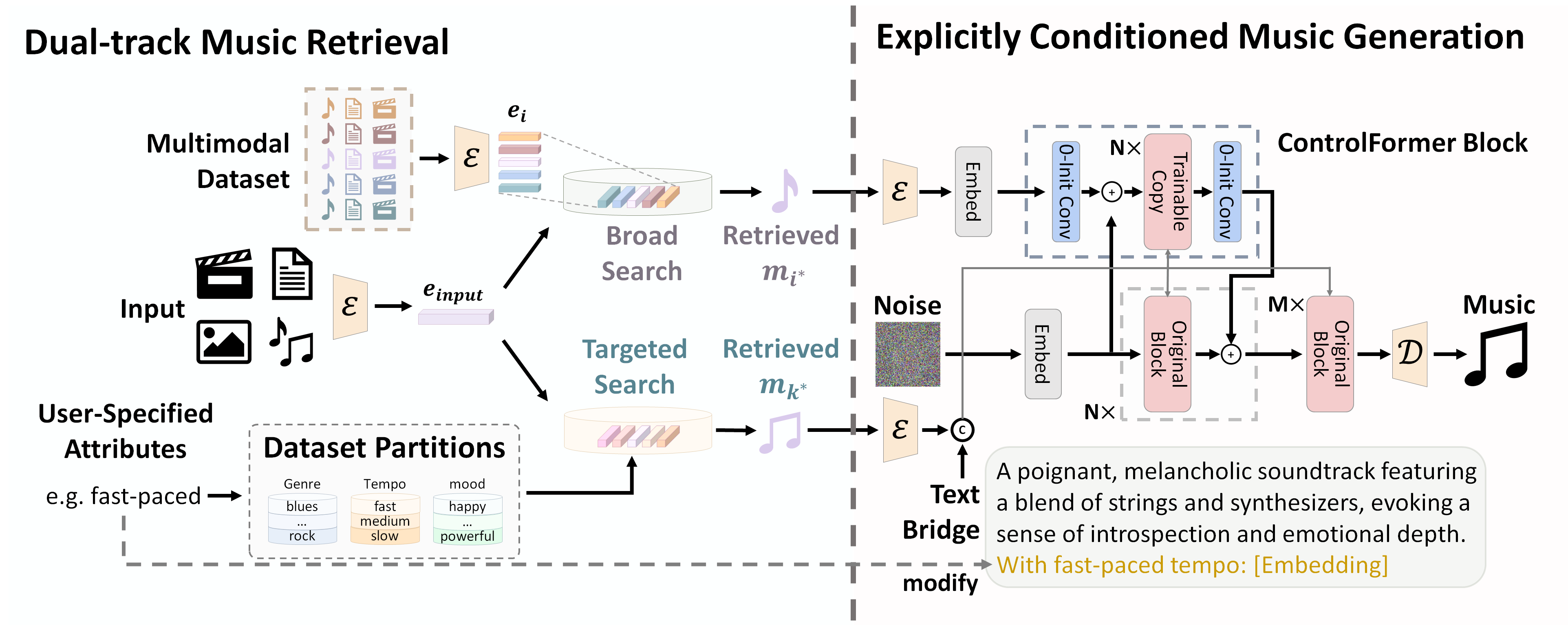}

   \vspace{-3mm}\caption{\textbf{Framework of Dual-track Music Retrieval and Explicitly Conditioned Music Generation.} The left part illustrates the Dual-track Music Retrieval process, which leverages our multimodal dataset to perform both broad and targeted retrieval. The right part shows the Explicitly Conditioned Music Generation pathway, where music is generated through a ControlFormer block integrating embeddings from selected music bridge, text bridge, and noisy inputs.}
   \vspace{-3mm}
   \label{fig:rag}
\end{figure*}

To bridge the gap between multimodal information and music modality, our approach employs textual descriptions as the first explicit bridge, effectively linking complex multimodal inputs with music generation. Leveraging the cross-modal understanding and generation ability of multimodal large language models~\cite{VLM:GPT-4v, mono_internvl, VLM:LLaVA-1.5}, we introduce the Multimodal Music Description Model (MMDM), which is built upon InternVL2~\cite{VLM:InternVL-1.5} to map multimodal signals into textual music description to guide the music generation process. 

The primary challenge is constructing effective video and music description paired data for efficient model training. To address this, we introduce a large-scale multimodal music dataset and detail the comprehensive dataset construction process, including data collection and the generation of music-description labels, which is shown in Fig.~\ref{fig:3.1}.

\noindent\textbf{\emph{Music Video Data Collection.}} In the data collection phase, we selected music videos (MVs) from online sources as primary dataset, focusing on those with a high degree of alignment between visual content and musical expression. Each entry in the dataset comprises three components:
\begin{itemize}
    \item Visual Content: Key frames were extracted from each MV to capture important scenes and transitions, establishing a coherent visual narrative.
    \item Audio Content: The original music tracks accompanying each MV serve as the primary auditory context. Using Demucs~\cite{demucs}, we isolate instrumental components by filtering out vocal parts to focus on the instrumental aspects.
    \item Metadata: We collect metadata such as titles, artist information, and video descriptions from platforms like YouTube and Shazam, providing supplementary context for each music-video pair.
\end{itemize}
    
\noindent\textbf{\emph{Automatic Music Description Generation.}}
After constructing the initial MV dataset, we proceed to generate music descriptions from the raw data. This process begins with an auto-tagging method that annotates each audio file across dimensions such as emotional tone and music theory elements. We then employ large language models to formulate unstructured metadata and tags into detailed music descriptions in natural language.

\underline{\emph{The auto-tagging process}} is grounded in a comprehensive label set developed by music experts, covering essential categories such as instruments, genres, and emotions for standardized and interpretable tagging. We extract the CLAP embeddings~\cite{clap} of candidate tags and align them with audio CLAP embeddings through cosine similarity:

\begin{equation} 
\text{cos}(\mathbf{e}_{\text{tag}}, \mathbf{e}_{\text{audio}}) = \frac{\mathbf{e}_{\text{tag}} \cdot \mathbf{e}_{\text{audio}}}{\| \mathbf{e}_{\text{tag}}\|  \|\mathbf{e}_{\text{audio}}\| } .
\end{equation} 

\noindent To ensure high precision in labeling, we use carefully calibrated similarity thresholds to filter out low-alignment tags.

\underline{\emph{To generate the detailed music descriptions}}, we combine auto-generated tags with crawled metadata to create comprehensive, context-rich descriptions. Leveraging the Llama 3.1 model~\cite{dubey2024llama31}, we craft descriptions that encapsulate the emotions and themes of each piece of music. By creating structured prompts, we guide the model to synthesize information from both tags and metadata, linking the music to relevant themes, emotions, and imagery. Through a series of carefully designed instructions, our approach produces nuanced descriptions that reflect the evocative and interpretative qualities of the music, providing a multifaceted understanding of each track.

\noindent\textbf{\emph{Multimodal Music Description Generation.}}
To use text as the bridge for multimodal music generation, we design a multimodal music description model that can extract visual and emotional cues from videos and images into textual music descriptions.
To incorporate visual and emotional cues for music description generation, we design an instruction system that guides the language model to first generate descriptions of the visual elements present in the input.
The MMDM is built by finetuning InternVL2~\cite{VLM:InternVL-1.5} on video-to-music description data. For efficient adaptation, we employed Low-Rank Adaptation (LoRA)~\cite{TL:LoRA}, enabling targeted fine-tuning while preserving the core multimodal strengths of InternVL2.

\subsection{Dual-track Music Retrieval}
\label{sec:dmr}

When musicians begin composing for multimedia, they often start by searching for existing music that aligns with the media's elements, learning from these examples to enhance their own compositions. Similarly, AI models can leverage retrieval-based techniques to assist in generation when datasets are insufficient, a process known as retrieval-augmented generation.

In the context of music generation, particularly with music descriptions obtained via MMDM, we aim to tackle the challenge of producing music that faithfully aligns with multimodal content. Direct text-to-music approaches often struggle to capture intricate musical elements such as chord progressions, tempo, and rhythm, revealing a gap between language and music modalities. To bridge this gap, we propose using music itself as a secondary explicit bridge. Inspired by the way composers select music to match the media and improve their compositions, we enhance the text-to-music generation model with a Dual-track Music Retrieval module. This module utilizes multimodal inputs to retrieve the most relevant music, which is then used as conditional input. This music-to-music conditioning enables a direct transfer of information, resulting in high-quality music outputs that closely reflect the intended multimodal content.

As illustrated in Fig.~\ref{fig:rag}, our dual-track music retrieval pipeline combines broad and targeted retrieval strategies for fine-grained and coarse-grained control, respectively. We leverage our auto-collected dataset in Sec.~\ref{sec:mmcm} as the retrieval database, containing 25,046 high-quality video-music-text triplets to provide external knowledge.

\noindent\textbf{\emph{Broad retrieval for Fine-Grained Control.}}
For each conditional modality, we first retrieve music that closely aligns with the input by computing similarities using CLIP~\cite{clip} or CLAP~\cite{clap} embeddings.
Given a target music, we directly use CLAP to compute audio embeddings $\mathbf{a}_{\text{input}}$ and compare them with embeddings $\{\mathbf{e}_i^{\text{audio}}\}$ in our music database: 
\begin{equation}
    i^* = \arg\max_{i} \cos\left( \mathbf{e}_{\text{input}}^{\text{audio}}, \mathbf{e}_i^{\text{audio}} \right).
\end{equation}
The music pieces retrieved from the broad retrieval $\mathbf{m}_{i^*}$ serve as fine-grained conditions, as illustrated in Sec.~\ref{sec:3.3}.
Notably, although we only apply music-centric retrieval during training, it is feasible to use CLIP embeddings of the input text $\mathbf{t}_{\text{input}}$ or visual signals $\mathbf{v}_{\text{input}}$ and compare them with text-music or visual-music pairs in the database as inference-time retrieval. The music associated with the closest matches is retrieved as:
\begin{equation}
    i^* = \arg\max_{i} \cos\left( \mathbf{e}_{\text{input}}^{\text{text/visual}}, \mathbf{e}_i^{\text{text/visual}} \right),
\end{equation}
where $\mathbf{e}_{\text{input}}^{\text{text/visual}}$ and $\mathbf{e}_i^{\text{text/visual}}$ are the CLIP embeddings of the input text or visual content and each dataset entry. This ensures that the retrieved music complements the semantic and emotional context of the visual or textual input, allowing the model to effectively capture complex musical features, such as melody and rhythm.

\begin{table*}[htbp] 
\centering 
\small
\renewcommand{\arraystretch}{1.1}
  \begin{tabular}{@{}lcccccccc@{}}
    \toprule
    \multirow{2}{*}{Method} &\multirow{2}{*}{Inference Time} & \multicolumn{3}{c}{Objective Metrics} & \multicolumn{4}{c}{Subjective Metrics$\uparrow$} \\
    \cmidrule(lr){3-5} \cmidrule(lr){6-9}
    && KL$_{passt}$$\downarrow$ & FD$_{openl3}$$\downarrow$ & IB$\uparrow$ & MP & EC & TC & RC \\
    \midrule
    CMT~\cite{cmt} & $\sim$~3 min  & 52.76 & 269.63 & 8.54  & 3.19 & 2.81 & 2.79 & 3.10 \\
    Video2music~\cite{video2music} & $\sim$~1 min  & 103.56 & 533.46 & 5.26 & 3.05 & 2.58 & 2.64 & 2.67 \\
    VidMuse~\cite{vidmuse} & $\sim$~13 min  & 56.48 & 187.13 & \textbf{22.09} & 3.01 & 2.91 & 3.05 & 3.02 \\
    M$^2$UGen~\cite{m2ugen} & $\sim$~40 s  & 60.41 & 180.72 & 15.58 & 2.84 & 2.32 & 2.37 & 2.71 \\
    VMB (ours) & $\sim$~20 s  & \textbf{48.84} & \textbf{105.84} & 21.62 & \textbf{3.85} & \textbf{3.36} & \textbf{3.38} & \textbf{3.62} \\
    \bottomrule
  \end{tabular}
\vspace{-2mm}
\caption{Video-to-music generation performance on SymMV dataset.  
 Up/down arrows indicate the desired direction for improvement.}
\label{tab:video2music} \vspace{-2mm}
\end{table*}

\noindent\textbf{\emph{Targeted retrieval for Coarse-Grained Control.}}
In parallel to the broad retrieval, we conduct a targeted retrieval within partitions of our dataset, specifically organized based on musical attributes labeled in Sec.~\ref{sec:mmcm}. For instance, the tempo partition is categorized into ``fast'', ``medium'', and ``slow'' subsets. This allows users to flexibly select songs that match specific attributes among genre, tempo, or mood partition. To retrieve a fast-paced song, we query directly within the ``fast'' subset in the tempo partition. In each subset, the most suitable music piece is determined by computing the cosine similarity between the CLAP embeddings of the desired textual attribute, encoded as $\mathbf{e}_{\text{desired}}^{\text{attr}}$ and $\{\mathbf{e}_k^{\text{audio}}\}$:
  \begin{equation}
    k^* = \arg\max_{k} \cos\left( \mathbf{e}_{\text{desired}}^{\text{attr}}, \mathbf{e}_k^{\text{audio}} \right).
\end{equation}
The embedding of the retrieved music $\mathbf{e}_{k^*}^{\text{audio}}$ is used as input in a subsequent module, where it is integrated into the generation process.   

\subsection{Explicitly Conditioned Music Generation}
\label{sec:3.3}
We first introduce our base music generation model, which leverages a latent diffusion transformer (DiT)~\cite{dit} to convert textual descriptions into music. Our model is built on the stable audio open framework~\cite{stableaudio}, where music is encoded into a compact latent space using a pre-trained VAE, with Gaussian noise added to the latent representation. A sequence of transformer blocks then processes the noisy input, with text features encoded by T5 and the current timestep injected into each block via cross-attention. The model is optimized using the standard diffusion objective with v-prediction~\cite{salimans2022progressive}.

\noindent\textbf{\emph{Music ControlFormer.}}
For retrieved music from broad retrieval, we draw inspiration from ControlNet~\cite{Controlnet, music_controlnet} since it is designed for structural guidance according to the conditions. However, considering original ControlNet is tailored for U-Net diffusion, we design Music ControlFormer for our diffusion transformer, which is shown in Fig.~\ref{fig:rag}. Rather than duplicating the entire model, we replicate only the early transformer layers of the model to create the ControlFormer branch. This selective duplication maintains computational efficiency while allowing control signals to influence the foundational stages of generation, where structural and semantic alignment can be established most effectively.

At each layer $l$, the main branch produces hidden states $\mathbf{h}_l^{\text{main}}$, and the ControlFormer produces $\mathbf{h}_l^{\text{control}}$. We combine these hidden states by element-wise addition:
\begin{equation} 
\tilde{\mathbf{h}}_l = \mathbf{h}_l^{\text{main}} + \mathbf{h}_l^{\text{control}}. 
\end{equation}

To ensure stable training and prevent disrupting the pretrained model's representations, we initialize the input and output convolution layers of the ControlFormer to zero, denoted as $\text{Conv}^{\mathcal{Z}}(\cdot)$. The hidden states are computed as:

\begin{equation} 
\mathbf{h}_l^{\text{control}} = \text{Conv}^{\mathcal{Z}}_l(\text{Block}_{1:l}^{\text{control}}(\text{Conv}^{\mathcal{Z}}_0({\mathbf{m}_{i^*}}))),
\end{equation}
where $\mathbf{m}_{i^*}$ represents the retrieved music embedding from broad retrieval and $\text{Encoder}_{1:l}^{\text{control}}$ indicates the sequential application of the first $l$ layers of the ControlFormer block.

\noindent\textbf{\emph{Stylization Module.}} Considering music from the targeted retrieval reflects specific user-specified attributes instead of fine-grained alignment, we introduce Stylization Module that effectively fuses the overall characteristic of the retrieved music with the generation process. We first add the textual description of the musical attribute to the end of the input prompt, which unlocks the ability to leverage language as an interface for guiding the overall style of the generated music. Then, the Stylization Module concatenates the CLAP embeddings of retrieved music with text and timestep embeddings to create a unified conditional representation. This combination is then adjusted to match the text embedding space's dimensions. We employ cross-attention to integrate the conditional representation into noisy music, to focus on stylistic cues from both the music and textual data. This method refines the alignment between the generated music and the specified attributes, enabling precise adjustments and yielding music that better fits the user's requirements.

\section{Experiments}

\begin{table*}[h]
\centering
\small
\renewcommand{\arraystretch}{1.2}
  \begin{tabular}{@{}lcccccc@{}}
    \toprule
    \multirow{2}{*}{Method} & \multicolumn{4}{c}{Objective Metrics} & \multicolumn{2}{c}{Subjective Metrics$\uparrow$} \\
    \cmidrule(lr){2-5} \cmidrule(lr){6-7}
    & KL$_{passt}$$\downarrow$ & FD$_{openl3}$$\downarrow$ & CLAPScore$\uparrow$ & IB$\uparrow$ & MP & TMA \\
    \midrule
    Stable Audio Open~\cite{stableaudio} &  42.89 & 183.09 & \textbf{40.92} & 24.67 & 3.41 & \textbf{3.52}\\
    MusicGen~\cite{musicgen} & 46.89 & 181.59 & 33.95 & 22.46 & 3.11 & 3.35 \\
    AudioLDM~\cite{audioldm} & 99.85 & 293.86 & 17.61 & 20.01 & 2.34 & 2.71 \\
    M$^2$UGen~\cite{m2ugen} & 49.03 & 188.84 & 28.76 & 16.70 & 3.19& 3.27 \\
    VMB (ours) & \textbf{37.43} & \textbf{132.16} & 39.66 & \textbf{29.36} & \textbf{3.78} & 3.48 \\
    \bottomrule
  \end{tabular}
\vspace{-2mm}
\caption{Text-to-music generation performance on SongDescriber dataset.}
\vspace{-2mm}
\label{tab:text2music}
\end{table*}

In this section, we conduct comprehensive evaluations of the proposed VMB framework on video-to-music (V2M), text-to-music (T2M), image-to-music (I2M) generation tasks, controllable generation, visuals-to-description generation, and ablation studies. We compare VMB with existing methods in zero-shot scenarios to ensure a fair assessment. Furthermore, we validate the effectiveness of the model's components through ablation studies.

\subsection{Experimental Setup}
\label{sec:exp_set}
\noindent\textbf{Model Training.} For \textbf{MMDM} in Sec.~\ref{sec:mmcm}, We use the InternVL2 model and AdamW optimizer with a learning rate of 1e-6. Training uses a local batch size of 2 with 8-step gradient accumulation. We adopt LoRA finetuning with the rank of 16 on 8 NVIDIA A800 GPUs trained for 10 epochs. During inference, we use top-k (k=50), nucleus sampling (p=1.0), and temperature 1.0. For \textbf{ECMG} in Sec.~\ref{sec:3.3}, We use a DiT model with T5 conditioning and an audio autoencoder for audio-text alignment. The model has 24 layers, 1536 dimensions, and 24 heads. Audio is processed with Oobleck encoders/decoders with a latent dim of 64. Training uses AdamW optimizer with a learning rate of 1e-6 and weight decay of 0.001. We conduct training on 4 NVIDIA A100 GPUs with a local batch size of 1 and 8-step gradient accumulation. More details are provided in Sec.~\ref{sec:implement}

\noindent\textbf{Training Dataset.} We use a dataset of 512K music tracks with text labels, collected using the labeling method described in Sec.~\ref{sec:mmcm}. We filtered these tracks by applying a PAM score~\cite{deshmukh2024pam} threshold, resulting in a subset of 54,112 high-quality tracks. Additionally, to support retrieval-augmented generation tasks, we incorporated a curated subset of dynamic music videos from the DISCO-200K-high-quality~\cite{disco10m}, along with videos collected by ourselves.

\noindent\textbf{Objective Metrics.} We used objective metrics including KL$_{passt}$~\cite{koutini2021efficient}, FD$_{openl3}$~\cite{cramer2019look}, CLAPScore~\cite{clap}, and ImageBind score (IB)~\cite{VLP:Imagebind}. 
KL$_{passt}$ and FD$_{openl3}$ evaluate the music quality from its statistical similarity to real music data and perceptual audio quality. CLAPScore measures the alignment between generated music descriptions and video content. IB measures the cross-modal semantic alignment between videos or images and the corresponding generated music. All objective metrics are scaled by 100.

\subsection{Video-to-music Generation}
We first evaluate the performance of our proposed VMB model on the task of video-to-music generation. Given an input video, models are expected to generate background music for the video with both music quality and video-music alignment. We use SymMV~\cite{musprod} as our evaluation set, which is a high-quality dataset specifically for video-to-music generation, featuring diverse video segments and music genres. We utilize all available videos from the SymMV dataset as of November 1, 2024. 
For consistency in evaluation across different methods, all video segments were truncated to 30 seconds and downsampled to 360p resolution. This configuration was selected to accommodate the requirements of certain models employed in our experiments. 

\noindent\textbf{Subjective Evaluation.} As a complementary to the objective evaluation, we follow previous methods~\cite{cmt, musprod, musicgen} to conduct subjective evaluation. We use metrics of Musical Pleasantness \textbf{(MP)}, Emotional Correspondence \textbf{(EC)}, Thematic Correspondence \textbf{(TC)}, Rhythmic Correspondence \textbf{(RC)}, and Text-Music Alignment \textbf{(TMA)}. Here EC, TC, and RC is only for the V2M task, while TMA is only for T2M. We generate background music for the same input video/text with different models, and send out questionnaires to ask participants to rate them in Likert Scale (from 1 to 5). We gathered a total of 64 valid responses. The questionnaire takes about 50 minutes to complete. Further details are provided in the supplementary material.

\noindent\textbf{Baselines.} We benchmark the performance of VMB against several state-of-the-art methods. For symbolic music generation, we include CMT~\cite{cmt} and Video2music~\cite{video2music}, which represent the current best in generating MIDI files from visual inputs. In the category of audio music generation, we compare against VidMuse~\cite{vidmuse} and M$^2$UGen~\cite{m2ugen}. VidMuse generates music tracks by progressively integrating visual cues, ensuring piecewise alignment with audiovisual elements, while M$^2$UGen utilizes large language models alongside a text-to-music model to transform visual inputs into musical outputs. 

\noindent\textbf{Results.} As shown in Tab.~\ref{tab:video2music}, VMB outperforms the baseline models on all the objective metrics. It achieves the lowest KL$_{passt}$ and FD$_{openl3}$ scores of 48.84 and 105.84, respectively, indicating that the music generated by VMB is statistically closer to real-world music and exhibits higher perceptual quality. Although its IB of 21.62 is slightly below VidMuse's 22.09, our model demonstrates effective semantic alignment between video and music. A notable feature is VMB's reduced inference time of about 20 seconds, significantly faster than VidMuse's 13 minutes. This improvement results from avoiding the extraction of redundant temporal features, while maintaining functionality by the retrieval module, making it ideal for applications requiring rapid music generation without compromising quality.

Subjective evaluations highlight its ability to create music that emotionally and thematically aligns with visuals. The DMR module ensures effective synchronization with video dynamics, enhancing the audiovisual narrative and user experience. While our method does not explicitly control local rhythm, subjective evaluations show significant improvements in rhythm metrics compared to baselines. We attribute this to the potential drawbacks of excessive focus on local rhythm at the cost of overall rhythm, which can disrupt coherence and emotional alignment with the visuals, limiting the music's impact on the narrative. We conduct an additional manual review to ensure thorough filtering. After this process, we retain a total of 279 text-music pairs.

\subsection{Text-to-music Generation}
We further evaluate the proposed VMB model on the text-to-music generation task. Given an input music description, each model is required to generate music that aligns with the description.
We report our results on the SongDescriber dataset~\cite{songdescriber}, a high-quality collection specifically curated for text-to-music tasks. We note that samples containing vocal parts are filtered out to ensure instrumental coherence.

\noindent\textbf{Baselines.} We compare VMB with several state-of-the-art text-to-music models. AudioLDM~\cite{audioldm} utilizes latent diffusion models for general audio tasks. MusicGen~\cite{musicgen} is a state-of-the-art transformer-based model trained on music-text pairs. Stable Audio Open~\cite{stableaudio} generates high-fidelity audio using a latent diffusion transformer conditioned on text. M$^2$UGen~\cite{m2ugen} also supports text-to-music generation through its utilization of large language models.

\noindent\textbf{Results.} As presented in Tab.~\ref{tab:text2music}, VMB outperforms baseline models across the objective metrics. It achieves the lowest KL$_{passt}$ of 37.43 and the lowest FD$_{openl3}$ of 132.16, indicating that the music generated by VMB is statistically closer to real-world music distributions and exhibits higher perceptual quality. VMB also attains a CLAPScore of 39.66, closely matching the ground truth of 40.34, highlighting its strong ability to capture the semantic content of text prompts. Additionally, VMB achieves the highest ImageBindScore of 29.36, further validating its robust performance in aligning text with music. Subjectively, VMB excels with the highest MP of 3.78, and TMA score of 3.48, which is slightly lower than the baseline's score of 3.52, suggesting it generates musically coherent and contextually accurate compositions.

The results demonstrate that VMB not only produces high-quality music but also aligns well with the given textual descriptions, outperforming existing state-of-the-art models in the text-to-music generation task.

\subsection{Image-to-music Generation}
To demonstrate the versatility of our proposed framework, we evaluated the VMB model on the image-to-music (I2M) generation task. In this setting, models are required to generate music that semantically and emotionally aligns with a given image. Evaluations were conducted using the MUImage~\cite{m2ugen} dataset, a high-quality collection of image-music pairs curated for nuanced cross-modal generation. We sort the entire dataset alphabetically by file name and used the first 1,500 image-music pairs as the test set. 

\noindent\textbf{Baselines.} We benchmarked VMB against state-of-the-art models, including CoDi~\cite{codi} and M$^2$UGen~\cite{m2ugen}. CoDi, an any-to-any generation model, is optimized for flexible cross-modal tasks, including I2M, making it a robust baseline for comparison.

\noindent\textbf{Results.} Tab.~\ref{tab:image2music} summarizes the performance of VMB, CoDi, and M$^2$UGen on the I2M task. VMB demonstrates a substantially lower KL$_{passt}$ and FD$_{openl3}$ compared to the baselines, alongside a higher Inception-Based Score (IB), indicating its ability to generate music that aligns closely with real-world distributions. Notably, although VMB is not explicitly trained on images, it effectively captures both the semantic and emotional content of visual inputs, underscoring its strong generalizability and its capacity to produce perceptually high-quality music from images.

\subsection{Controllable Generation}
To assess the controllability of our model, we evaluate its capability to generate music with distinct attributes, specifically focusing on genre, instrument, mood, and tempo. We select contrasting attribute pairs, such as ``happy'' vs. ``non-happy'' moods, and random sample 20 songs for each attribute pair. For each song, we generated 10 variations conditioned on the sampled song itself to assess the model's capability to adjust each attribute independently. For each attribute, we report the average change in CLAPScore ($\Delta$), quantifying how well the generated music aligns with the specified textual description.

Tab.~\ref{table:control} summarizes the average $\Delta$ in CLAPScore across instrument, genre, and mood controls. Instrument control achieves the largest average change, indicating that the model effectively differentiates between instruments, likely due to CLAP's sensitivity to instrument features. Genre and mood control yield smaller $\Delta$ values, suggesting that the model can modulate genre and emotional tone. We also conduct additional experiment in Sec.~\ref{sec:control_supp}
The results highlight the VMB's ability for attribute-specific control, demonstrating its adaptability to diverse inputs and effective modulation of musical attributes.
\begin{table}[t]
\small
\centering
\renewcommand{\arraystretch}{1.2}
  \begin{tabular}{@{}lccc@{}}
    \toprule
    Method & KL$_{passt}$$\downarrow$ & FD$_{openl3}$$\downarrow$ & IB$\uparrow$\\
    \midrule
    CoDi~\cite{codi} & 216.48 & 251.52 & 9.60 \\
    M$^2$UGen~\cite{m2ugen} & 128.33 & 247.42 & 2.28 \\
    VMB (ours) & \textbf{105.60} & \textbf{119.76} & \textbf{11.88} \\
    \bottomrule
  \end{tabular}
\vspace{-2mm}
\caption{Image-to-music generation performance on MUImage dataset.}
\vspace{-2mm}
\label{tab:image2music}
\end{table}

\subsection{Visual-to-Description Generation}
We evaluated our MMDM model on a subset of 8,042 dynamic videos from the DISCO-200K-high-quality dataset~\cite{disco10m}, selected by filtering out static or lyrics-based videos through pixel difference analysis and manual review.
To assess the quality of the generated descriptions, we used the CLAPScore to measure the similarity between the original and generated textual descriptions in terms of their alignment with the corresponding music.

As shown in Tab.~\ref{table:v2d}, our MMDM model achieves a CLAPScore of 50.88, outperforming the baseline models GPT-4V and InternVL, which scored 44.41 and 44.21, respectively.  GPT-4V~\cite{gpt4v} is a SOTA multimodal large language model (MLLM) that extends the capabilities of GPT-4~\cite{VLM:GPT-4} to handle both image and text inputs. While MLLMs excel at generating descriptive text from visual cues, GPT-4V and InternVL are not specifically designed for music-text generation tasks, which may account for their lower performance in this context. 

These results confirm the superior description generation capability of the MMDM model, demonstrating its ability to produce more accurate and contextually relevant descriptions of music compared to the baselines.
\begin{table}[]
\begin{minipage}[t]{0.48\linewidth}
\centering

\vspace{1mm}
\renewcommand\arraystretch{1.2}
\resizebox{0.9\linewidth}{!}{
    \begin{tabular}{lc}
    \toprule
    Model &  CLAPScore\\ 
    \midrule
    GPT-4V~\cite{gpt4v} & 44.41  \\
    InternVL~\cite{VLM:InternVL-1.5} & 44.21 \\
    MMDM& 50.88 \\
    \bottomrule
    \end{tabular}
}
\vspace{-2mm}
\caption{Video-to-Description Generation Performance.}
\label{table:v2d}
\end{minipage}
\hspace{1.5mm}
\begin{minipage}[t]{0.48\linewidth}
\centering
\vspace{1mm}
\renewcommand\arraystretch{1.0}
\resizebox{0.9\linewidth}{!}{
    \begin{tabular}{lc}
    \toprule
    Attribute &  Change ($\Delta$)\\ 
    \midrule
    Instrument & +11.46  \\
    Genre &  +3.03\\
    Mood & +4.14 \\
    \bottomrule
    \end{tabular}
}
\vspace{-2mm}
\caption{Attribute control effectiveness measured by average change ($\Delta$) in CLAPScore.}
\label{table:control}
\end{minipage}
\vspace{-0.8em}
\end{table}

\subsection{Ablation Studies}

We conduct ablation studies to assess the individual contributions of two key components in our framework: broad retrieval (BR) and targeted retrieval (TR). The ablations are performed on the video-to-music generation task using the SymMV dataset.

The results of these ablations are presented in Tab.~\ref{tab:ablation}, demonstrate that combining broad retrieval (BR) and targeted retrieval (TR) yields the best performance across all metrics. Specifically, the combination achieves the lowest KL$_{passt}$ (75.29) and FD$_{openl3}$ (177.27), indicating improved thematic alignment and perceptual quality. BR improves alignment with the input content, while TR enhances creativity, reflected in the higher IB (24.70). These findings highlight the complementary roles of BR and TR in improving music generation.

When either broad or targeted retrieval is omitted, the performance decreases, particularly in terms of KL$_{passt}$ and FD$_{openl3}$, further emphasizing the complementary nature of these components. The ablation results demonstrate that both broad and targeted retrieval are crucial for achieving high-quality, relevant, and creative music generation in the video-to-music task.

\begin{table}[t]
\small
  \centering
  \renewcommand{\arraystretch}{1.2}
  \begin{tabular}{cc|ccccc}
    \toprule
    BR & TR & KL$_{passt}$$\downarrow$ & FD$_{openl3}$$\downarrow$ & IB$\uparrow$ \\
    \midrule
    $\checkmark$ & $\checkmark$ &  \textbf{75.29} & \textbf{177.27} & \textbf{24.70}\\
    $\checkmark$ & $\times$     & 91.89 & 199.74 & 20.73 \\
    $\times$ & $\checkmark$& 91.07 & 387.14 & 20.51 \\
    $\times$ & $\times$& 96.42 & 360.29 & 14.67 \\
    \bottomrule
  \end{tabular}
  \caption{Ablation of model components on video-to-music generation with SymMV dataset. BR, TR represent broad retrieval and target retrieval respectively.}
  \label{tab:ablation}
\end{table}

\section{Broader Impacts, Limitations, and Future Works}

\textbf{Broader Impacts.} 
The VMB framework introduces several significant impacts across various domains. 
Primarily, VMB enhances accessibility in entertainment technologies such as gaming and virtual reality by enabling the automatic generation of emotionally resonant background music, which could improve user engagement and accessibility for people with diverse abilities. 
However, the technology also presents potential negative impacts. 
The automation of music generation might reduce the need for human composers in certain contexts, potentially affecting their livelihoods. Moreover, biases in the training data could lead to outputs that do not adequately represent the diversity of global musical expressions, potentially introducing unfairness in musical representation.

\textbf{Limitations.} The VMB framework represents a significant advancement in multimodal music generation; however, it is not devoid of limitations. The model's effectiveness heavily relies on the diversity and quality of the dataset it is trained on. Presently, available datasets may not adequately represent the vast array of musical styles and cultural expressions, thus limiting the system's ability to produce a broad spectrum of musical outputs. Moreover, the challenge of accurately translating complex emotional and thematic nuances across different modalities persists, as the system occasionally fails to capture the depth and subtlety inherent in human compositions.

\textbf{Future Works.} Future enhancements to the VMB framework should focus on several key areas. Broadening the diversity of the dataset to encompass a greater variety of musical styles and cultural expressions would considerably enhance the model's generative capabilities. Improving the system's understanding of and ability to accurately translate intricate emotional and thematic nuances between modalities would make the technology more effective and emotionally impactful. 
Integrating music theory into the music generation process could provide a more robust framework for generating musically coherent outputs.

\section{Conclusion}

In this paper, we propose a novel multi-modal music generation system named VMB to address the challenges of existing methods. VMB accepts diverse inputs, including text, image, and video, and integrates them effectively by using text and music as a bridge.  Our system enables fine-grained control over key musical elements, enabling users to direct the generation process according to their preferences.
We conduct extensive experiments to demonstrate that VMB generates music that aligns well with multimodal inputs and exhibits high controllability, outperforming current state-of-the-art methods. VMB holds significant potential for multimedia applications, facilitating personalized and contextually rich music generation across domains such as entertainment and interactive media.

{
    \small
    \bibliographystyle{ieeenat_fullname}
    \bibliography{main}

\begin{thebibliography}{61}
\providecommand{\natexlab}[1]{#1}
\providecommand{\url}[1]{\texttt{#1}}
\expandafter\ifx\csname urlstyle\endcsname\relax
  \providecommand{\doi}[1]{doi: #1}\else
  \providecommand{\doi}{doi: \begingroup \urlstyle{rm}\Url}\fi

\bibitem[Agostinelli et~al.(2023)Agostinelli, Denk, Borsos, Engel, Verzetti, Caillon, Huang, Jansen, Roberts, Tagliasacchi, et~al.]{musiclm}
Andrea Agostinelli, Timo~I Denk, Zal{\'a}n Borsos, Jesse Engel, Mauro Verzetti, Antoine Caillon, Qingqing Huang, Aren Jansen, Adam Roberts, Marco Tagliasacchi, et~al.
\newblock Musiclm: Generating music from text.
\newblock \emph{arXiv preprint arXiv:2301.11325}, 2023.

\bibitem[Aguilar~Rangel et~al.(2022)Aguilar~Rangel, Bedwell, Costanzi, Taylor, Russo, Bernardes, Ricagno, Frydman, Vendruscolo, and Sormanni]{aguilar2022fragment}
Mauricio Aguilar~Rangel, Alice Bedwell, Elisa Costanzi, Ross~J Taylor, Rosaria Russo, Gon{\c{c}}alo~JL Bernardes, Stefano Ricagno, Judith Frydman, Michele Vendruscolo, and Pietro Sormanni.
\newblock Fragment-based computational design of antibodies targeting structured epitopes.
\newblock \emph{Science Advances}, 8\penalty0 (45):\penalty0 eabp9540, 2022.

\bibitem[Caffagni et~al.(2024)Caffagni, Cocchi, Moratelli, Sarto, Cornia, Baraldi, and Cucchiara]{caffagni2024wiki}
Davide Caffagni, Federico Cocchi, Nicholas Moratelli, Sara Sarto, Marcella Cornia, Lorenzo Baraldi, and Rita Cucchiara.
\newblock Wiki-llava: Hierarchical retrieval-augmented generation for multimodal llms.
\newblock In \emph{Proceedings of the IEEE/CVF Conference on Computer Vision and Pattern Recognition}, pages 1818--1826, 2024.

\bibitem[Chen et~al.(2022)Chen, Hu, Saharia, and Cohen]{chen2022re}
Wenhu Chen, Hexiang Hu, Chitwan Saharia, and William~W Cohen.
\newblock Re-imagen: Retrieval-augmented text-to-image generator.
\newblock \emph{arXiv preprint arXiv:2209.14491}, 2022.

\bibitem[Chen et~al.(2023)Chen, Wu, Wang, Su, Chen, Xing, Zhong, Zhang, Zhu, Lu, Li, Luo, Lu, Qiao, and Dai]{VLM:InternVL}
Zhe Chen, Jiannan Wu, Wenhai Wang, Weijie Su, Guo Chen, Sen Xing, Muyan Zhong, Qinglong Zhang, Xizhou Zhu, Lewei Lu, Bin Li, Ping Luo, Tong Lu, Yu Qiao, and Jifeng Dai.
\newblock Internvl: Scaling up vision foundation models and aligning for generic visual-linguistic tasks.
\newblock \emph{arXiv: 2312.14238}, 2023.

\bibitem[Chen et~al.(2024)Chen, Wang, Tian, Ye, Gao, Cui, Tong, Hu, Luo, Ma, et~al.]{VLM:InternVL-1.5}
Zhe Chen, Weiyun Wang, Hao Tian, Shenglong Ye, Zhangwei Gao, Erfei Cui, Wenwen Tong, Kongzhi Hu, Jiapeng Luo, Zheng Ma, et~al.
\newblock How far are we to gpt-4v? closing the gap to commercial multimodal models with open-source suites.
\newblock \emph{arXiv:2404.16821}, 2024.

\bibitem[Chowdhury et~al.(2024)Chowdhury, Nag, Joseph, Srinivasan, and Manocha]{chowdhury2024melfusion}
Sanjoy Chowdhury, Sayan Nag, KJ Joseph, Balaji~Vasan Srinivasan, and Dinesh Manocha.
\newblock Melfusion: Synthesizing music from image and language cues using diffusion models.
\newblock In \emph{Proceedings of the IEEE/CVF Conference on Computer Vision and Pattern Recognition}, pages 26826--26835, 2024.

\bibitem[Copet et~al.(2024)Copet, Kreuk, Gat, Remez, Kant, Synnaeve, Adi, and D{\'e}fossez]{musicgen}
Jade Copet, Felix Kreuk, Itai Gat, Tal Remez, David Kant, Gabriel Synnaeve, Yossi Adi, and Alexandre D{\'e}fossez.
\newblock Simple and controllable music generation.
\newblock \emph{Advances in Neural Information Processing Systems}, 36, 2024.

\bibitem[Cramer et~al.(2019)Cramer, Wu, Salamon, and Bello]{cramer2019look}
Aurora~Linh Cramer, Ho-Hsiang Wu, Justin Salamon, and Juan~Pablo Bello.
\newblock Look, listen, and learn more: Design choices for deep audio embeddings.
\newblock In \emph{ICASSP}, pages 3852--3856. IEEE, 2019.

\bibitem[Dao(2023)]{flashattention_v2}
Tri Dao.
\newblock Flashattention-2: Faster attention with better parallelism and work partitioning.
\newblock \emph{arXiv preprint arXiv:2307.08691}, 2023.

\bibitem[D{\'e}fossez et~al.(2019)D{\'e}fossez, Usunier, Bottou, and Bach]{demucs}
Alexandre D{\'e}fossez, Nicolas Usunier, L{\'e}on Bottou, and Francis Bach.
\newblock Demucs: Deep extractor for music sources with extra unlabeled data remixed.
\newblock \emph{arXiv preprint arXiv:1909.01174}, 2019.

\bibitem[Deshmukh et~al.(2024)Deshmukh, Alharthi, Elizalde, Gamper, Ismail, Singh, Raj, and Wang]{deshmukh2024pam}
Soham Deshmukh, Dareen Alharthi, Benjamin Elizalde, Hannes Gamper, Mahmoud~Al Ismail, Rita Singh, Bhiksha Raj, and Huaming Wang.
\newblock Pam: Prompting audio-language models for audio quality assessment.
\newblock \emph{arXiv preprint arXiv:2402.00282}, 2024.

\bibitem[Di et~al.(2021)Di, Jiang, Liu, Wang, Zhu, He, Liu, and Yan]{cmt}
Shangzhe Di, Zeren Jiang, Si Liu, Zhaokai Wang, Leyan Zhu, Zexin He, Hongming Liu, and Shuicheng Yan.
\newblock Video background music generation with controllable music transformer.
\newblock In \emph{MM}, 2021.

\bibitem[Dong et~al.(2018)Dong, Hsiao, Yang, and Yang]{musegan}
Hao-Wen Dong, Wen-Yi Hsiao, Li-Chia Yang, and Yi-Hsuan Yang.
\newblock Musegan: Multi-track sequential generative adversarial networks for symbolic music generation and accompaniment.
\newblock In \emph{AAAI}, 2018.

\bibitem[Dubey et~al.(2024)Dubey, Jauhri, Pandey, Kadian, Al-Dahle, Letman, Mathur, Schelten, Yang, Fan, et~al.]{dubey2024llama31}
Abhimanyu Dubey, Abhinav Jauhri, Abhinav Pandey, Abhishek Kadian, Ahmad Al-Dahle, Aiesha Letman, Akhil Mathur, Alan Schelten, Amy Yang, Angela Fan, et~al.
\newblock The llama 3 herd of models.
\newblock \emph{arXiv preprint arXiv:2407.21783}, 2024.

\bibitem[Elizalde et~al.(2023)Elizalde, Deshmukh, Al~Ismail, and Wang]{clap}
Benjamin Elizalde, Soham Deshmukh, Mahmoud Al~Ismail, and Huaming Wang.
\newblock Clap learning audio concepts from natural language supervision.
\newblock In \emph{ICASSP 2023-2023 IEEE International Conference on Acoustics, Speech and Signal Processing (ICASSP)}, pages 1--5. IEEE, 2023.

\bibitem[Evans et~al.(2024)Evans, Parker, Carr, Zukowski, Taylor, and Pons]{stableaudio}
Zach Evans, Julian~D Parker, CJ Carr, Zack Zukowski, Josiah Taylor, and Jordi Pons.
\newblock Stable audio open.
\newblock \emph{arXiv preprint arXiv:2407.14358}, 2024.

\bibitem[Girdhar et~al.(2023)Girdhar, El{-}Nouby, Liu, Singh, Alwala, Joulin, and Misra]{VLP:Imagebind}
Rohit Girdhar, Alaaeldin El{-}Nouby, Zhuang Liu, Mannat Singh, Kalyan~Vasudev Alwala, Armand Joulin, and Ishan Misra.
\newblock Imagebind one embedding space to bind them all.
\newblock In \emph{CVPR}, pages 15180--15190, 2023.

\bibitem[Guu et~al.(2020)Guu, Lee, Tung, Pasupat, and Chang]{guu2020retrieval}
Kelvin Guu, Kenton Lee, Zora Tung, Panupong Pasupat, and Mingwei Chang.
\newblock Retrieval augmented language model pre-training.
\newblock In \emph{International conference on machine learning}, pages 3929--3938. PMLR, 2020.

\bibitem[Hu et~al.(2022)Hu, Shen, Wallis, Allen{-}Zhu, Li, Wang, Wang, and Chen]{TL:LoRA}
Edward~J. Hu, Yelong Shen, Phillip Wallis, Zeyuan Allen{-}Zhu, Yuanzhi Li, Shean Wang, Lu Wang, and Weizhu Chen.
\newblock Lora: Low-rank adaptation of large language models.
\newblock In \emph{ICLR}, 2022.

\bibitem[Hu et~al.(2023)Hu, Iscen, Sun, Wang, Chang, Sun, Schmid, Ross, and Fathi]{hu2023reveal}
Ziniu Hu, Ahmet Iscen, Chen Sun, Zirui Wang, Kai-Wei Chang, Yizhou Sun, Cordelia Schmid, David~A Ross, and Alireza Fathi.
\newblock Reveal: Retrieval-augmented visual-language pre-training with multi-source multimodal knowledge memory.
\newblock In \emph{Proceedings of the IEEE/CVF conference on computer vision and pattern recognition}, pages 23369--23379, 2023.

\bibitem[Huang et~al.(2019)Huang, Vaswani, Uszkoreit, Simon, Hawthorne, Shazeer, Dai, Hoffman, Dinculescu, and Eck]{music_transformer}
Cheng-Zhi~Anna Huang, Ashish Vaswani, Jakob Uszkoreit, Ian Simon, Curtis Hawthorne, Noam Shazeer, Andrew~M Dai, Matthew~D Hoffman, Monica Dinculescu, and Douglas Eck.
\newblock Music transformer: Generating music with long-term structure.
\newblock In \emph{ICLR}, 2019.

\bibitem[Huang et~al.(2023)Huang, Park, Wang, Denk, Ly, Chen, Zhang, Zhang, Yu, Frank, et~al.]{noise2music}
Qingqing Huang, Daniel~S Park, Tao Wang, Timo~I Denk, Andy Ly, Nanxin Chen, Zhengdong Zhang, Zhishuai Zhang, Jiahui Yu, Christian Frank, et~al.
\newblock Noise2music: Text-conditioned music generation with diffusion models.
\newblock \emph{arXiv preprint arXiv:2302.03917}, 2023.

\bibitem[Huang and Yang(2020)]{remi}
Yu-Siang Huang and Yi-Hsuan Yang.
\newblock Pop music transformer: Beat-based modeling and generation of expressive pop piano compositions.
\newblock In \emph{MM}, 2020.

\bibitem[Jang et~al.(2023)Jang, Yang, Zhang, Jin, and Chowdhury]{jang2023oobleck}
Insu Jang, Zhenning Yang, Zhen Zhang, Xin Jin, and Mosharaf Chowdhury.
\newblock Oobleck: Resilient distributed training of large models using pipeline templates.
\newblock In \emph{Proceedings of the 29th Symposium on Operating Systems Principles}, pages 382--395, 2023.

\bibitem[Kang et~al.(2024)Kang, Poria, and Herremans]{video2music}
Jaeyong Kang, Soujanya Poria, and Dorien Herremans.
\newblock Video2music: Suitable music generation from videos using an affective multimodal transformer model.
\newblock \emph{Expert Systems with Applications}, page 123640, 2024.

\bibitem[Kong et~al.(2020)Kong, Cao, Iqbal, Wang, Wang, and Plumbley]{kong2020panns}
Qiuqiang Kong, Yin Cao, Turab Iqbal, Yuxuan Wang, Wenwu Wang, and Mark~D Plumbley.
\newblock Panns: Large-scale pretrained audio neural networks for audio pattern recognition.
\newblock \emph{IEEE/ACM Transactions on Audio, Speech, and Language Processing}, 28:\penalty0 2880--2894, 2020.

\bibitem[Koutini et~al.(2021)Koutini, Schl{\"u}ter, Eghbal-Zadeh, and Widmer]{koutini2021efficient}
Khaled Koutini, Jan Schl{\"u}ter, Hamid Eghbal-Zadeh, and Gerhard Widmer.
\newblock Efficient training of audio transformers with patchout.
\newblock \emph{arXiv preprint arXiv:2110.05069}, 2021.

\bibitem[Lanzend{\"o}rfer et~al.(2024)Lanzend{\"o}rfer, Gr{\"o}tschla, Funke, and Wattenhofer]{disco10m}
Luca Lanzend{\"o}rfer, Florian Gr{\"o}tschla, Emil Funke, and Roger Wattenhofer.
\newblock Disco-10m: A large-scale music dataset.
\newblock \emph{Advances in Neural Information Processing Systems}, 36, 2024.

\bibitem[Lewis et~al.(2020)Lewis, Perez, Piktus, Petroni, Karpukhin, Goyal, K{\"u}ttler, Lewis, Yih, Rockt{\"a}schel, et~al.]{lewis2020retrieval}
Patrick Lewis, Ethan Perez, Aleksandra Piktus, Fabio Petroni, Vladimir Karpukhin, Naman Goyal, Heinrich K{\"u}ttler, Mike Lewis, Wen-tau Yih, Tim Rockt{\"a}schel, et~al.
\newblock Retrieval-augmented generation for knowledge-intensive nlp tasks.
\newblock \emph{Advances in Neural Information Processing Systems}, 33:\penalty0 9459--9474, 2020.

\bibitem[Li et~al.(2024)Li, Qin, Zheng, Jin, and Liu]{diffbgm}
Sizhe Li, Yiming Qin, Minghang Zheng, Xin Jin, and Yang Liu.
\newblock Diff-bgm: A diffusion model for video background music generation.
\newblock In \emph{Proceedings of the IEEE/CVF Conference on Computer Vision and Pattern Recognition}, pages 27348--27357, 2024.

\bibitem[Liu et~al.(2023{\natexlab{a}})Liu, Li, Li, and Lee]{VLM:LLaVA-1.5}
Haotian Liu, Chunyuan Li, Yuheng Li, and Yong~Jae Lee.
\newblock Improved baselines with visual instruction tuning.
\newblock \emph{arXiv: 2310.03744}, 2023{\natexlab{a}}.

\bibitem[Liu et~al.(2024)Liu, Yuan, Liu, Mei, Kong, Tian, Wang, Wang, Wang, and Plumbley]{audioldm}
Haohe Liu, Yi Yuan, Xubo Liu, Xinhao Mei, Qiuqiang Kong, Qiao Tian, Yuping Wang, Wenwu Wang, Yuxuan Wang, and Mark~D Plumbley.
\newblock Audioldm 2: Learning holistic audio generation with self-supervised pretraining.
\newblock \emph{IEEE/ACM Transactions on Audio, Speech, and Language Processing}, 2024.

\bibitem[Liu et~al.(2023{\natexlab{b}})Liu, Hussain, Sun, and Shan]{m2ugen}
Shansong Liu, Atin~Sakkeer Hussain, Chenshuo Sun, and Ying Shan.
\newblock M$^{2}$ugen: Multi-modal music understanding and generation with the power of large language models.
\newblock \emph{arXiv preprint arXiv:2311.11255}, 2023{\natexlab{b}}.

\bibitem[Long et~al.(2022)Long, Yin, Ajanthan, Nguyen, Purkait, Garg, Blair, Shen, and van~den Hengel]{long2022retrieval}
Alexander Long, Wei Yin, Thalaiyasingam Ajanthan, Vu Nguyen, Pulak Purkait, Ravi Garg, Alan Blair, Chunhua Shen, and Anton van~den Hengel.
\newblock Retrieval augmented classification for long-tail visual recognition.
\newblock In \emph{CVPR}, pages 6959--6969, 2022.

\bibitem[Loshchilov(2017)]{adamw}
I Loshchilov.
\newblock Decoupled weight decay regularization.
\newblock \emph{arXiv preprint arXiv:1711.05101}, 2017.

\bibitem[Lu et~al.(2022)Lu, Zhou, Bao, Chen, Li, and Zhu]{lu2022dpm}
Cheng Lu, Yuhao Zhou, Fan Bao, Jianfei Chen, Chongxuan Li, and Jun Zhu.
\newblock Dpm-solver++: Fast solver for guided sampling of diffusion probabilistic models.
\newblock \emph{arXiv preprint arXiv:2211.01095}, 2022.

\bibitem[Luo et~al.(2024)Luo, Yang, Dou, Wang, Dai, Qiao, and Zhu]{mono_internvl}
Gen Luo, Xue Yang, Wenhan Dou, Zhaokai Wang, Jifeng Dai, Yu Qiao, and Xizhou Zhu.
\newblock Mono-internvl: Pushing the boundaries of monolithic multimodal large language models with endogenous visual pre-training.
\newblock \emph{arXiv preprint arXiv:2410.08202}, 2024.

\bibitem[Manco et~al.(2023)Manco, Weck, Doh, Won, Zhang, Bogdanov, Wu, Chen, Tovstogan, Benetos, et~al.]{songdescriber}
Ilaria Manco, Benno Weck, Seungheon Doh, Minz Won, Yixiao Zhang, Dmitry Bogdanov, Yusong Wu, Ke Chen, Philip Tovstogan, Emmanouil Benetos, et~al.
\newblock The song describer dataset: a corpus of audio captions for music-and-language evaluation.
\newblock \emph{arXiv preprint arXiv:2311.10057}, 2023.

\bibitem[OpenAI(2023{\natexlab{a}})]{VLM:GPT-4}
OpenAI.
\newblock {GPT-4} technical report.
\newblock \emph{arXiv: 2303.08774}, 2023{\natexlab{a}}.

\bibitem[OpenAI(2023{\natexlab{b}})]{gpt4v}
OpenAI.
\newblock Gpt-4v(ision) system card.
\newblock 2023{\natexlab{b}}.

\bibitem[Peebles and Xie(2023)]{dit}
William Peebles and Saining Xie.
\newblock Scalable diffusion models with transformers.
\newblock In \emph{CVPR}, pages 4195--4205, 2023.

\bibitem[Radford et~al.(2021)Radford, Kim, Hallacy, Ramesh, Goh, Agarwal, Sastry, Askell, Mishkin, Clark, et~al.]{clip}
Alec Radford, Jong~Wook Kim, Chris Hallacy, Aditya Ramesh, Gabriel Goh, Sandhini Agarwal, Girish Sastry, Amanda Askell, Pamela Mishkin, Jack Clark, et~al.
\newblock Learning transferable visual models from natural language supervision.
\newblock In \emph{ICML}, 2021.

\bibitem[Raffel et~al.(2020)Raffel, Shazeer, Roberts, Lee, Narang, Matena, Zhou, Li, and Liu]{TransF:T5}
Colin Raffel, Noam Shazeer, Adam Roberts, Katherine Lee, Sharan Narang, Michael Matena, Yanqi Zhou, Wei Li, and Peter~J. Liu.
\newblock Exploring the limits of transfer learning with a unified text-to-text transformer.
\newblock \emph{JMLR}, 21:\penalty0 140:1--140:67, 2020.

\bibitem[Rajbhandari et~al.(2020)Rajbhandari, Rasley, Ruwase, and He]{rajbhandari2020zero}
Samyam Rajbhandari, Jeff Rasley, Olatunji Ruwase, and Yuxiong He.
\newblock Zero: Memory optimizations toward training trillion parameter models.
\newblock In \emph{SC20: International Conference for High Performance Computing, Networking, Storage and Analysis}, pages 1--16. IEEE, 2020.

\bibitem[Roberts et~al.(2018)Roberts, Engel, Raffel, Hawthorne, and Eck]{musicvae}
Adam Roberts, Jesse Engel, Colin Raffel, Curtis Hawthorne, and Douglas Eck.
\newblock A hierarchical latent vector model for learning long-term structure in music.
\newblock In \emph{ICML}, 2018.

\bibitem[Salimans and Ho(2022)]{salimans2022progressive}
Tim Salimans and Jonathan Ho.
\newblock Progressive distillation for fast sampling of diffusion models.
\newblock \emph{arXiv preprint arXiv:2202.00512}, 2022.

\bibitem[Sheynin et~al.(2022)Sheynin, Ashual, Polyak, Singer, Gafni, Nachmani, and Taigman]{sheynin2022knn}
Shelly Sheynin, Oron Ashual, Adam Polyak, Uriel Singer, Oran Gafni, Eliya Nachmani, and Yaniv Taigman.
\newblock Knn-diffusion: Image generation via large-scale retrieval.
\newblock \emph{arXiv preprint arXiv:2204.02849}, 2022.

\bibitem[Su et~al.(2023)Su, Li, Huang, Kuzmin, Lee, Donahue, Sha, Jansen, Wang, Verzetti, et~al.]{v2meow}
Kun Su, Judith~Yue Li, Qingqing Huang, Dima Kuzmin, Joonseok Lee, Chris Donahue, Fei Sha, Aren Jansen, Yu Wang, Mauro Verzetti, et~al.
\newblock V2meow: Meowing to the visual beat via music generation.
\newblock \emph{arXiv preprint arXiv:2305.06594}, 2023.

\bibitem[Tang et~al.(2024)Tang, Yang, Zhu, Zeng, and Bansal]{codi}
Zineng Tang, Ziyi Yang, Chenguang Zhu, Michael Zeng, and Mohit Bansal.
\newblock Any-to-any generation via composable diffusion.
\newblock \emph{NeurIPS}, 36, 2024.

\bibitem[Tian et~al.(2024)Tian, Liu, Yuan, Pan, Huang, Liu, Tan, Chen, Xue, and Guo]{vidmuse}
Zeyue Tian, Zhaoyang Liu, Ruibin Yuan, Jiahao Pan, Xiaoqiang Huang, Qifeng Liu, Xu Tan, Qifeng Chen, Wei Xue, and Yike Guo.
\newblock Vidmuse: A simple video-to-music generation framework with long-short-term modeling.
\newblock \emph{arXiv preprint arXiv:2406.04321}, 2024.

\bibitem[Wang et~al.(2023)Wang, Chen, and Li]{wang2023continuous}
Yajie Wang, Mulin Chen, and Xuelong Li.
\newblock Continuous emotion-based image-to-music generation.
\newblock \emph{IEEE Transactions on Multimedia}, 2023.

\bibitem[Wang et~al.(2024)Wang, Wang, Hancke, Liu, and Lau]{wang2024themestation}
Zhenwei Wang, Tengfei Wang, Gerhard Hancke, Ziwei Liu, and Rynson~WH Lau.
\newblock Themestation: Generating theme-aware 3d assets from few exemplars.
\newblock In \emph{ACM SIGGRAPH 2024 Conference Papers}, pages 1--12, 2024.

\bibitem[Wu and Zheng(2022)]{wu2022learning}
Rundi Wu and Changxi Zheng.
\newblock Learning to generate 3d shapes from a single example.
\newblock \emph{arXiv preprint arXiv:2208.02946}, 2022.

\bibitem[Wu et~al.(2023)Wu, Fei, Qu, Ji, and Chua]{VLM:NeXTGPT}
Shengqiong Wu, Hao Fei, Leigang Qu, Wei Ji, and Tat-Seng Chua.
\newblock Next-gpt: Any-to-any multimodal llm.
\newblock \emph{arXiv: 2309.05519}, 2023.

\bibitem[Wu et~al.(2024)Wu, Donahue, Watanabe, and Bryan]{music_controlnet}
Shih-Lun Wu, Chris Donahue, Shinji Watanabe, and Nicholas~J Bryan.
\newblock Music controlnet: Multiple time-varying controls for music generation.
\newblock \emph{IEEE/ACM Transactions on Audio, Speech, and Language Processing}, 32:\penalty0 2692--2703, 2024.

\bibitem[Yang et~al.(2017)Yang, Chou, and Yang]{midinet}
Li-Chia Yang, Szu-Yu Chou, and Yi-Hsuan Yang.
\newblock Midinet: A convolutional generative adversarial network for symbolic-domain music generation.
\newblock In \emph{ISMIR}, 2017.

\bibitem[Yang et~al.(2023)Yang, Li, Lin, Wang, Lin, Liu, and Wang]{VLM:GPT-4v}
Zhengyuan Yang, Linjie Li, Kevin Lin, Jianfeng Wang, Chung-Ching Lin, Zicheng Liu, and Lijuan Wang.
\newblock The dawn of lmms: Preliminary explorations with gpt-4v (ision).
\newblock \emph{arXiv: 2309.17421}, 9, 2023.

\bibitem[Zhang et~al.(2023{\natexlab{a}})Zhang, Rao, and Agrawala]{Controlnet}
Lvmin Zhang, Anyi Rao, and Maneesh Agrawala.
\newblock Adding conditional control to text-to-image diffusion models, 2023{\natexlab{a}}.

\bibitem[Zhang et~al.(2023{\natexlab{b}})Zhang, Xiao, Liu, Dou, and Nie]{zhang2023retrieve}
Peitian Zhang, Shitao Xiao, Zheng Liu, Zhicheng Dou, and Jian-Yun Nie.
\newblock Retrieve anything to augment large language models.
\newblock \emph{arXiv preprint arXiv:2310.07554}, 2023{\natexlab{b}}.

\bibitem[Zhuo et~al.(2023)Zhuo, Wang, Wang, Liao, Bao, Peng, Han, Zhang, Fang, and Liu]{musprod}
Le Zhuo, Zhaokai Wang, Baisen Wang, Yue Liao, Chenxi Bao, Stanley Peng, Songhao Han, Aixi Zhang, Fei Fang, and Si Liu.
\newblock Video background music generation: Dataset, method and evaluation.
\newblock In \emph{Proceedings of the IEEE/CVF International Conference on Computer Vision}, pages 15637--15647, 2023.

\end{thebibliography}
}
\clearpage

\appendix
\section{Dataset Analysis}

We utilize four self-curated datasets for the following purposes: 1) training the ECMG module; 2) constructing the DMR retrieval dataset; 3) conducting subjective and qualitative evaluations; and 4) assessing the performance of the MMDM module. Additionally, we document modifications made to existing datasets, namely SymMV~\cite{musprod} and SongDescriber~\cite{songdescriber}.

\subsection{ECMG Training Dataset}
\label{sec:sup_train_ds}
\begin{figure}[t]
  \centering
   \includegraphics[width=\linewidth]{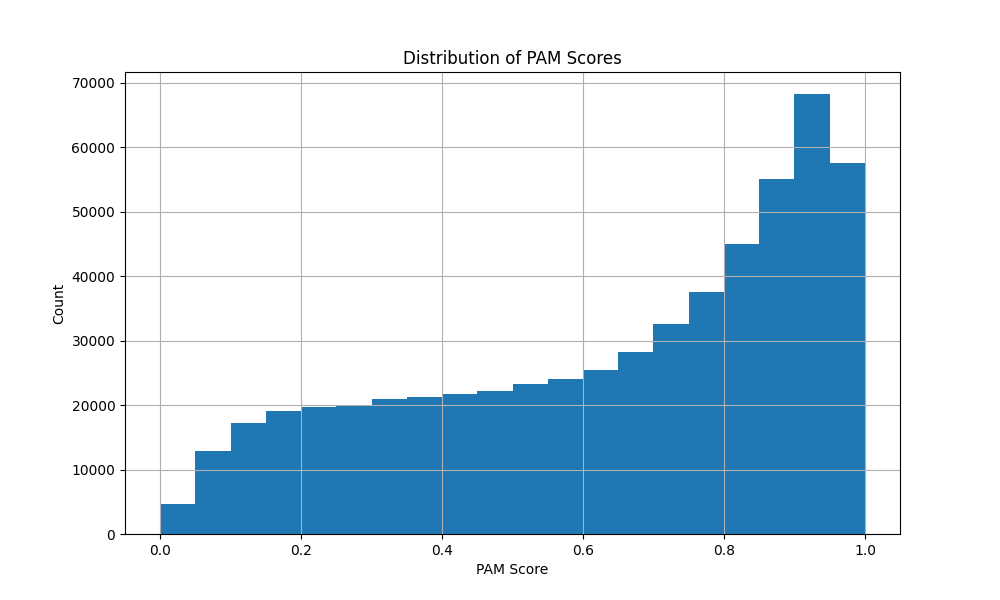}

   \vspace{-3mm}\caption{Distribution of PAM Scores across the raw training dataset.}
   \vspace{-3mm}
   \label{fig:pam}
\end{figure}

As outlined in Sec.~\ref{sec:exp_set}, we collect a total of 512K music tracks. Tracks containing vocal components are identified using PANN~\cite{kong2020panns} and CLAP~\cite{clap}, based on pre-defined tags such as \texttt{vocal}, \texttt{choir}, and \texttt{human voice}. Scores are computed between each track and the target tags, and a threshold of 0.1 is selected to identify tracks with vocal components. This threshold is determined through a manual review process, where we listen to a subset of music tracks to ensure the balance between accurately identifying vocal parts and minimizing false positives. The decision to remove tracks with human voices follows previous works~\cite{vidmuse}, and is guided by three key considerations: 1) to lower the complexity of modeling voice components; 2) to enable the model to focus more effectively on instrumental music; and 3) to align with the primary goal of video background music generation~\cite{cmt, musprod}, which typically excludes vocal elements. Additionally, we filter out low-quality tracks by applying the PAM~\cite{deshmukh2024pam} score with a threshold of 0.95. After filtering, we obtain 54,112 high-quality tracks. The distribution of PAM score in the whole dataset is illustrated in Fig.~\ref{fig:pam}.

The selected tracks are annotated using the methodology described in Sec.~\ref{sec:mmcm}. Candidate tags are pre-defined by musicians who serve as domain experts to align with commonly used labels in music generation while minimizing the effect of long-tail distribution. The detailed annotations are provided in the file \texttt{label.json}. 

We use \textbf{instrument} tagging as an example to illustrate the experts' taxonomy. Instruments are first categorized into five primary groups: strings, keyboards, wind instruments, percussion instruments, and others. To address the issue of long-tail distributions, rare tags are combined into broader categories. For instance, instruments like the double bass, which are seldom played independently, are treated as indivisible units. Therefore, we use strings to replace double bass. This approach ensures robust categorization while accurately reflecting the practical usage of instruments in compositions. Other tagging principles, which align with the general design philosophy of instrument categorization, are omitted here for brevity. 

To generate natural descriptions, we employ the Llama-3.1-8B-Instruct~\cite{dubey2024llama31} model to generate one-sentence descriptions for each music track. The system prompt is: \textit{You are a professional music expert. Here are tags about this music. You need to generate a one-sentence description for this music. Only generate the description without any other information.} Metadata tags were supplied via the user prompt.

\begin{figure}[t]
  \centering
   \includegraphics[width=0.9\linewidth]{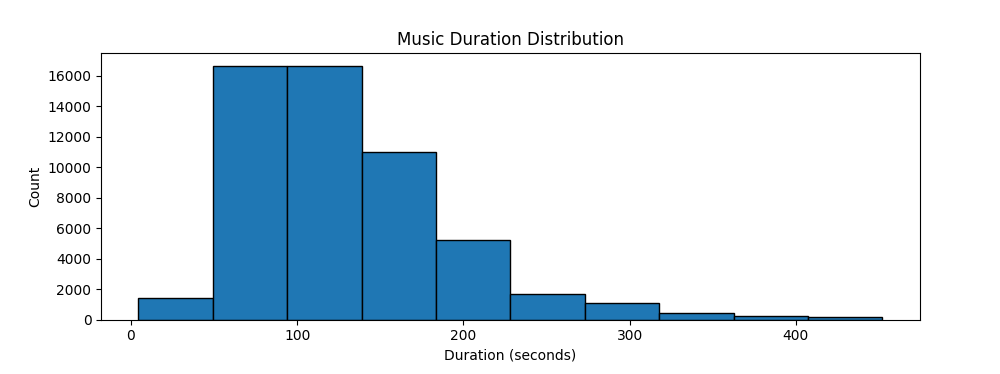}

   \vspace{-3mm}\caption{Histogram of music duration in the training dataset.}
   \vspace{-3mm}
   \label{fig:trainingdata1}
\end{figure}

\begin{figure}[t]
  \centering
   \includegraphics[width=0.9\linewidth]{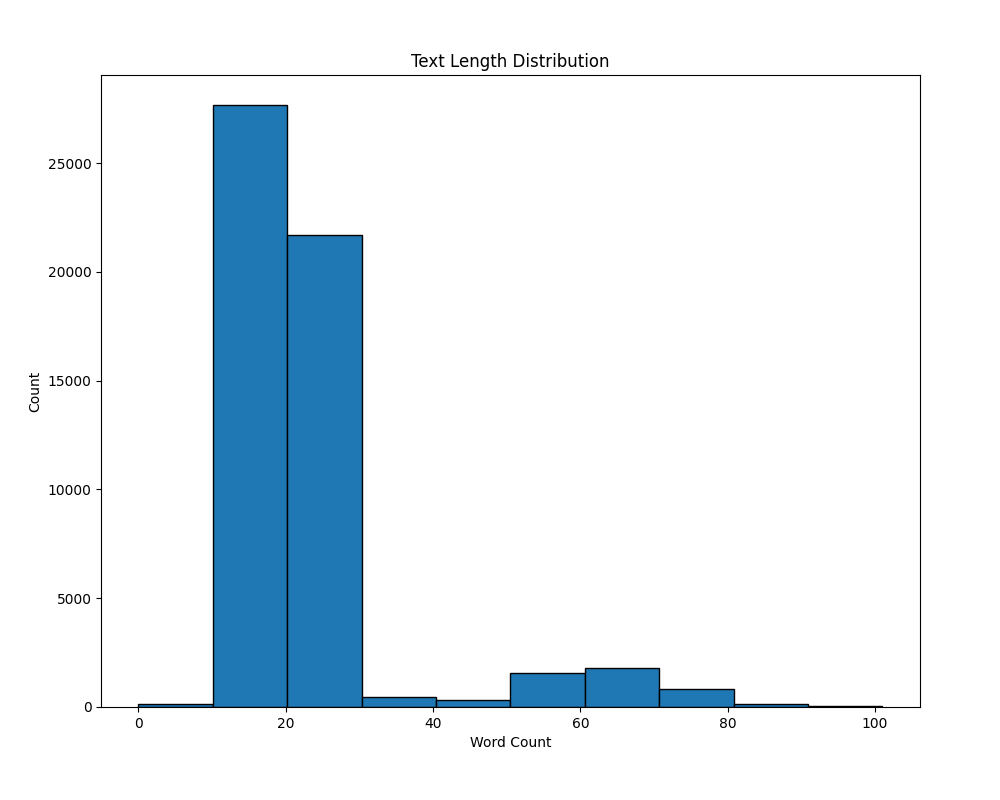}

   \vspace{-3mm}\caption{Histogram of text word counts in the training dataset.}
   \vspace{-3mm}
   \label{fig:trainingdata2}
\end{figure}

Fig.~\ref{fig:trainingdata1} and Fig.~\ref{fig:trainingdata2} illustrate the distribution of music durations and the corresponding descriptions, providing insight into the characteristics of the dataset.

\subsection{MMDM Training and Retrieval Dataset}
\label{sec:mmdmset}
We have introduced the methodology of collecting the MMDM training and retrieval dataset in Sec. ~\ref{sec:mmcm} Here, we provide other details.

First, we scrape the hottest singers and their songs from Spotify. We use YoutubeAPI\footnote{https://developers.google.com/youtube/v3} to search for their original official music video with keywords in the format of \texttt{\{singer\} + \{song name\} + Official Music Video}. We scrape all the music videos with their Youtube description boxes and top-10 comments. We also use Shazam\footnote{https://www.shazam.com/} to get other metadata. Finally we obtain the primary dataset introduced in Sec. ~\ref{sec:mmcm}

We use methodology stated in Sec. ~\ref{sec:mmcm} and Sec.~\ref{sec:sup_train_ds} to label each music track. Eventually our dataset contains 24,719 video-text-music pairs.

\begin{figure}[t]
  \centering
   \includegraphics[width=0.9\linewidth]{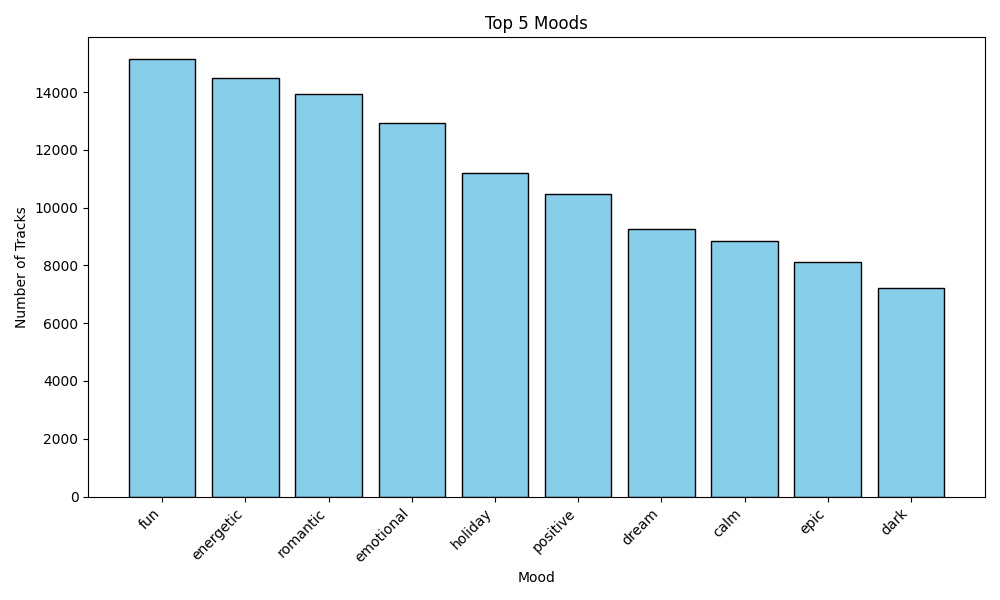}
   \caption{Distribution of mood tags across the retrieval dataset. This histogram shows the frequency of various mood categories, illustrating the emotional diversity captured in our data.}
   \label{fig:mood}
\end{figure}

\begin{figure}[t]
  \centering
   \includegraphics[width=0.9\linewidth]{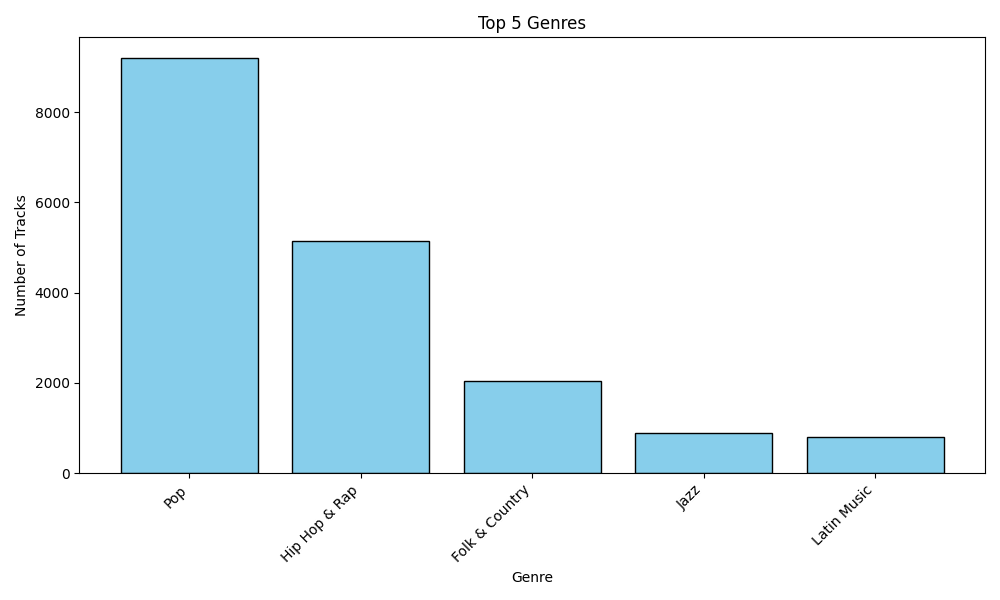}
   \caption{Genre distribution within the retrieval dataset. This bar graph reflects the variety of music genres represented, indicating the dataset's broad applicability for genre-specific retrieval tasks.}
   \label{fig:genre}
\end{figure}

\begin{figure}[t]
  \centering
   \includegraphics[width=0.9\linewidth]{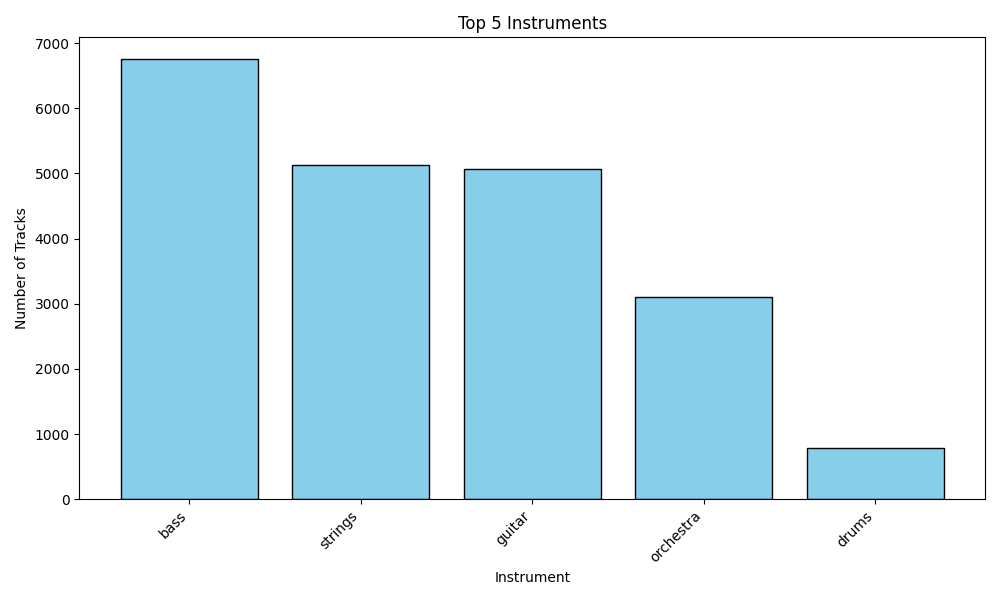}
   \caption{Histogram of instrument tags in our retrieval dataset. This figure shows the range of musical instruments represented, underscoring the dataset's comprehensive coverage of instrumental music.}
   \label{fig:instrument}
\end{figure}

\begin{figure}[t]
  \centering
   \includegraphics[width=0.9\linewidth]{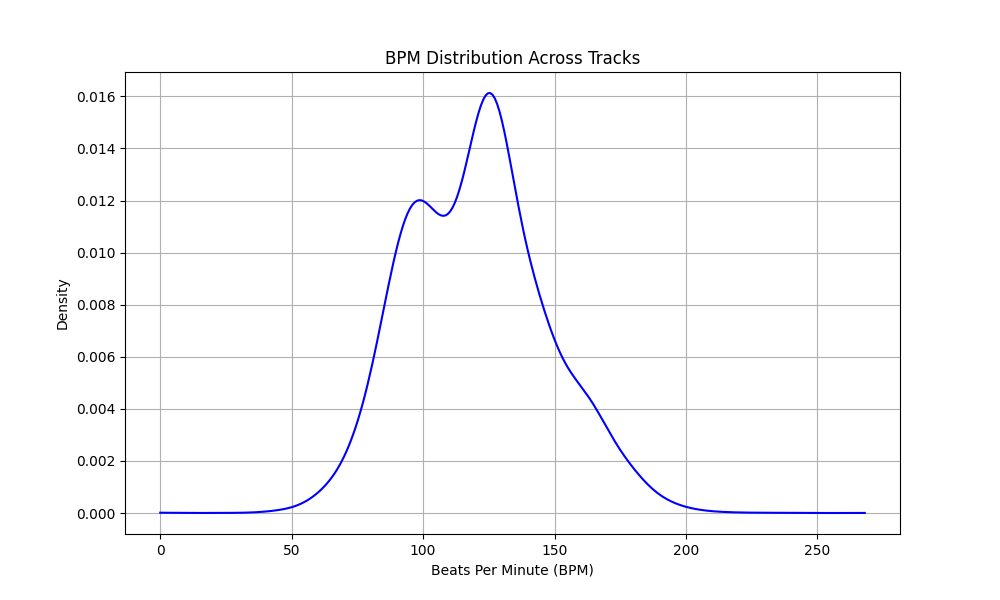}
   \caption{Density curve of BPM across the retrieval dataset. This plot illustrates the distribution of Beats Per Minute, showcasing the tempo range covered in our collection.}
   \label{fig:bpm}
\end{figure}

In targeted retrieval detailed introduced in Sec.~\ref{sec:dmr}, we partition our whole datasets. We follow three standards and get three different partitions, \emph{i.e.}, genre, tempo, and mood partition. We use the labeled tags to partition the whole dataset. Distribution of each attribute is shown in Fig.~\ref{fig:mood}, ~\ref{fig:genre}, ~\ref{fig:instrument}, and ~\ref{fig:bpm}.

\begin{figure}[t]
  \centering
   \includegraphics[width=0.9\linewidth]{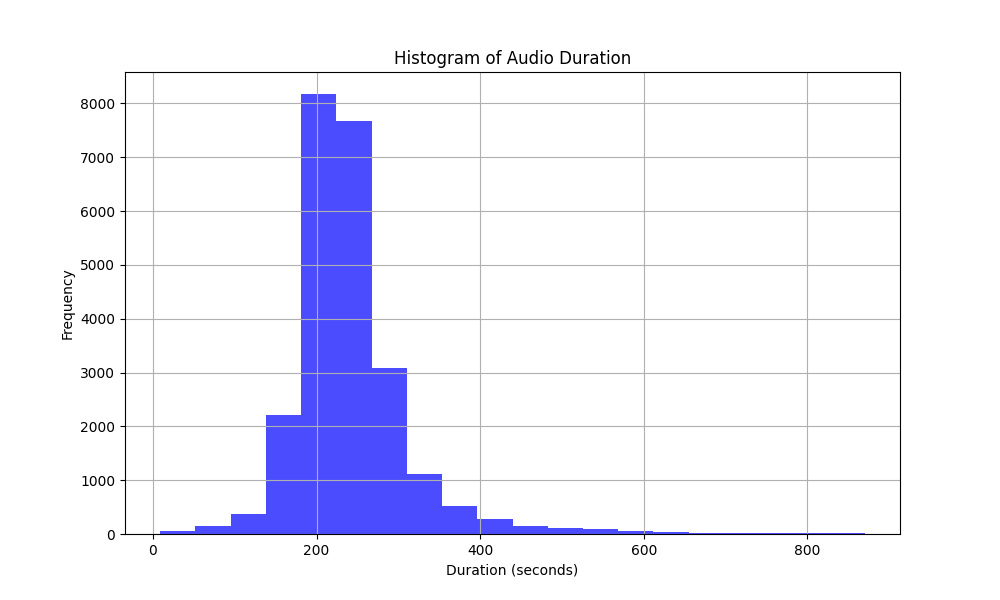}

   \vspace{-3mm}\caption{Histogram of audio durations in retrieval dataset. This shows the distribution of song lengths in the dataset.}
   \vspace{-3mm}
   \label{fig:audio_duration}
\end{figure}

\begin{figure}[t]
  \centering
   \includegraphics[width=0.9\linewidth]{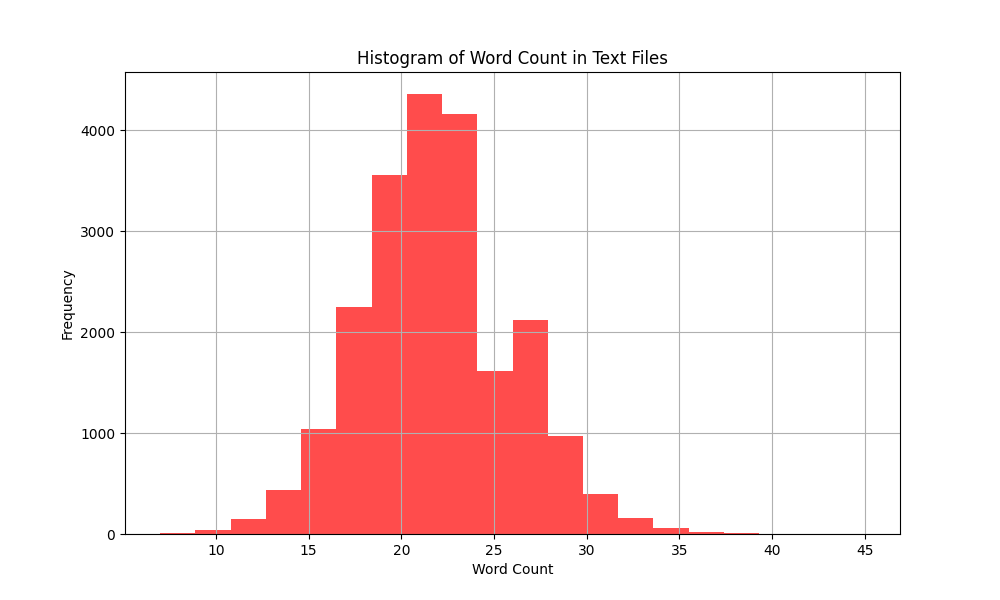}

   \vspace{-3mm}\caption{Histogram of text word counts in retrieval dataset. This represents the distribution of word counts in the associated text data.}
   \vspace{-3mm}
   \label{fig:word_count}
\end{figure}

\begin{figure}[t]
  \centering
   \includegraphics[width=0.9\linewidth]{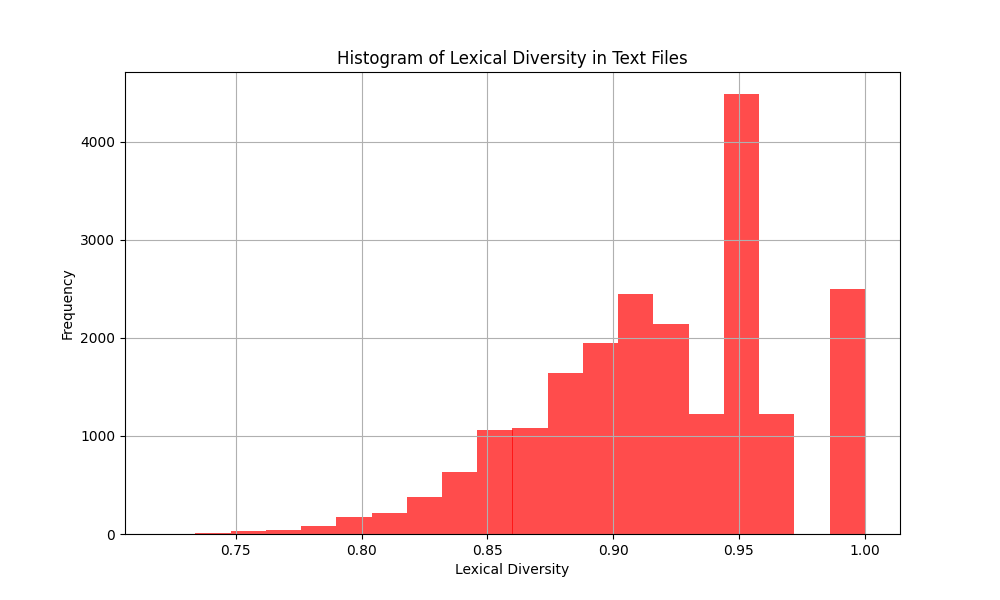}

   \vspace{-3mm}\caption{Histogram of lexical diversity scores in retrieval dataset. This shows the variation in vocabulary usage across text samples.}
   \vspace{-3mm}
   \label{fig:lexical_diversity}
\end{figure}

Fig.~\ref{fig:audio_duration} shows the distribution of audio durations in the dataset. The majority of the audio files have durations concentrated within a specific range, suggesting consistent lengths across the dataset. The associated text data is characterized by word count distribution in Fig.~\ref{fig:word_count}, showing a variety of text lengths, with most falling in the mid-range. Finally, the lexical diversity, representing vocabulary usage variation, is displayed in Fig.~\ref{fig:lexical_diversity}. Most texts demonstrate high lexical diversity, indicating rich vocabulary usage across the dataset.

\subsection{Subjective Evaluation Dataset}

To evaluate our model across diverse video contexts, we construct a dataset encompassing seven distinct video categories: Scene/Vlog, Documentary, Advertisement, Movie, Game, Anime, and Sports. These categories are selected to ensure a comprehensive benchmark that reflects a wide range of video types and their corresponding music needs. Each video is manually reviewed to ensure quality, resulting in a final dataset of 35 high-quality videos.

We apply the following criteria for video selection:
\begin{itemize}
    \item Duration: Videos are limited to a maximum length of three minutes to ensure efficient evaluation.
    \item Content Quality: Only classic or high-quality videos are included, reflecting diverse and impactful visual themes.
\end{itemize}

\subsection{Visuals-to-Description Evaluation Dataset}

We downloaded all available videos from the Disco-200K-high-quality dataset~\cite{disco10m} as of July 30, 2023. Upon review, we noted that many videos consisted of static images or were lyrics-based music videos. To filter for dynamic content, we calculated the average pixel difference between consecutive frames to identify static videos. Videos close to the threshold arere manually reviewed to confirm their dynamic nature. Following this rigorous selection process, we curated a subset of 8,042 videos that demonstrated sufficient motion suitable for our analysis.

\subsection{SongDescriber and MUImage}
\label{sec:modification}

While Stable Audio Open~\cite{stableaudio} claimed to have filtered the SongDescriber dataset to exclude tracks with vocal components, 

For the MUImage dataset~\cite{m2ugen}, as the authors do not specify a valid test set but only provided the training set, we sort the entire dataset alphabetically by file name and used the first 1,500 image-music pairs as the test set. Consequently, the evaluation of M$^2$UGen on MUImage is actually conducted on its training set.

The specific video-music / text-music pairs used in our study are documented in the files \texttt{SymMV.csv} and \texttt{SongDescriber.csv}.

\begin{figure*}[!t]
    \centering
    \includegraphics[width=\linewidth]{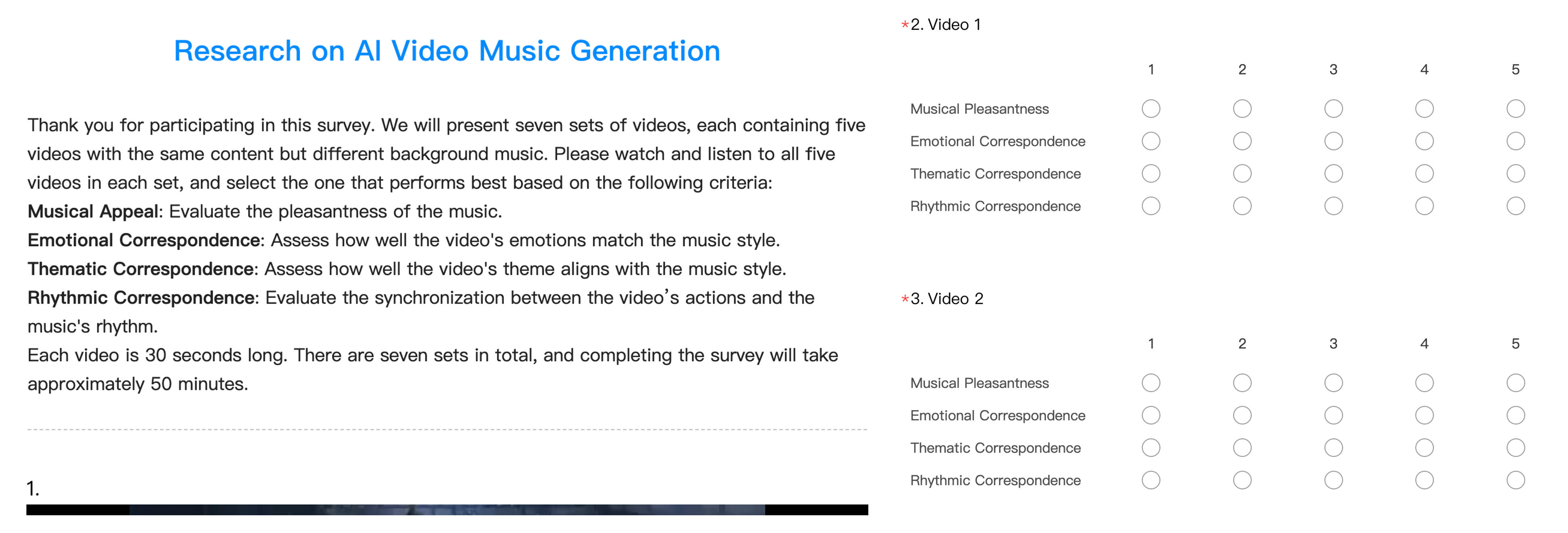}
    \caption{Screenshot of the user study questionnaire in subjective evaluation.}
    \label{fig:userstudy}
    \vspace{-2mm}
\end{figure*}

\section{Experiment Details}

In this section, we provide details of the experiments in the main paper.

\subsection{Implementation Details}
\label{sec:implement}
For MMDM module stated in Sec. ~\ref{sec:mmcm}, We utilize the InternVL2~\cite{VLM:InternVL-1.5} architecture with mixed precision training (bfloat16) and the AdamW optimizer (learning rate: 1e-6, cosine decay). Top-k sampling (k=50) and nucleus sampling (p=1.0) are used for generation, with a temperature of 1.0. The model features FlashAttention v2~\cite{flashattention_v2}, 32 layers, and 16 attention heads, paired with a vision backbone (Intern-ViT-6B)~\cite{VLM:InternVL} at 448x448 resolution. Training is performed using a batch size of 2 with gradient accumulation over 8 steps. Fine-tuning is conducted with LoRA~\cite{TL:LoRA} at rank 16, utilizing 8 NVIDIA A800-SXM4-80GB GPUs over 3 days for a total of 10 epochs.

In ECMG module introduced in Sec.~\ref{sec:3.3}, We leverage a DiT~\cite{dit} model with T5~\cite{TransF:T5} conditioning and an audio autoencoder for fine-tuning on audio-text alignment tasks. The architecture includes a 24-layer continuous transformer with 1536 embedding dimensions and 24 attention heads. Audio inputs are preprocessed via an autoencoder featuring Oobleck~\cite{jang2023oobleck} encoders/decoders with multi-scale channel configurations and a latent dimension of 64.

Training uses AdamW~\cite{adamw} (learning rate: 1e-6, weight decay: 0.001) with an InverseLR scheduler. The model operates at a 44.1kHz sample rate with a batch size of 1 and gradient accumulation over 8 steps. Mixed precision (16-bit) training is employed, using DeepSpeed~\cite{rajbhandari2020zero} for optimization with 4 NVIDIA A100-PCIE-40GB GPUs.

Conditioning involves audio features, time attributes, and textual prompts (via T5-base, max length: 128). 

\subsection{Video-to-Music Generation}
We evaluate video-to-music generation using several state-of-the-art models, adhering to their default settings for fair comparison. Due to the sequence length limitation of M$^2$UGen~\cite{m2ugen}, all samples are truncated to 30 seconds. Additionally, to accommodate the resolution constraint of CMT~\cite{cmt}, all videos are resized to 360p.

We follow the default setting in each generation model. For VMB, we set a standard sample rate of 44,100 Hz and adjust the CFG scale to 7 for balancing conditioning influence, with 100 steps for detailed noise reduction. Our sampler, ``DPM++ 3M SDE''~\cite{lu2022dpm}, manages noise sampling across a sigma range from 0.03 to 1000.

In our experiments, we use a server with dual Intel Xeon Processors (Icelake) with 64 cores each and four NVIDIA A100-PCIE-40GB GPUs, under a KVM virtualized Linux environment. These configurations ensure the reproducibility of our inference time measurements.

\subsection{Text-to-Music Generation}
For text-to-music models, we also adhere to their default settings for a fair comparison. Due to the maximum sequence length limitation of M$^2$UGen~\cite{m2ugen}, all generated music samples are capped at 30 seconds. 

VMB follows the parameters outlined in~\cite{stableaudio}. It uses a 44,100 Hz sample rate, a CFG scale of 7, 100 steps for diffusion, and a ``DPM++ 3M SDE'' sampler with a sigma range from 0.03 to 1000.

\begin{table}[]
\centering
\vspace{-2mm}
\caption{Average BPM of music generated under varying tempo conditions.}
\resizebox{0.5\linewidth}{!}{
    \begin{tabular}{lc}
    \toprule
    Model &  Average BPM\\ 
    \midrule
    Fast & 143.55  \\
    Medium & 122.64 \\
    Slow& 93.88 \\
    \bottomrule
    \end{tabular}
}
\label{tab:bpm}
\end{table}

\subsection{Subjective Evaluation}

For the subjective evaluation, we provide each participant with two sets of evaluations: one for video-to-music generation and another for text-to-music generation. Each set consists of seven groups, corresponding to the seven categories in our subjective evaluation dataset. For each category, a video is randomly sampled. Each group include five music pieces, each generated by one of the models.

For video-to-music (V2M) generation, participants are briefed on the purpose of the study and instructed to evaluate how well the generated music matched the video in terms of mood, thematic alignment, and overall enhancement of the viewing experience. Each video is presented seven times, each time paired with a background music track generated from a different model. The order of music presentations is randomized to eliminate order effects. After watching all versions of a video, participants are asked to rate each version using a Likert scale, \emph{i.e.} rate from 1 to 5, considering factors such as emotional impact, thematic coherence, and suitability for the video's content. Additionally, participants are encouraged to provide qualitative feedback explaining their preferences, highlighting specific emotional or thematic factors that influenced their choices.

For the V2M task, we utilize the following metrics to guide participant evaluations:
\begin{itemize}
    \item \textbf{Musical Pleasantness (MP)}: The aesthetic quality of the music, independent of context.
    \item \textbf{Emotional Correspondence (EC)}: How well the music conveys the intended emotions of the video.
    \item \textbf{Thematic Correspondence (TC)}: The alignment between the video's theme and the generated music.
    \item \textbf{Rhythmic Correspondence (RC)}: The synchronization between the video's motion and the music's rhythm.
\end{itemize}

The evaluation process for text-to-music (T2M) generation follows a similar structure. Participants are provided with textual prompts and asked to evaluate the generated music based on how well it reflects the mood, style, and thematic elements described in the text. Each textual prompt is paired with five music tracks generated from five models, and participants rate the tracks using the same Likert scoring system. Feedback is collected to identify specific strengths and weaknesses of the generated music in conveying textual meaning.

For the T2M task, we select the following metrics:
\begin{itemize}
    \item \textbf{Musical Pleasantness (MP)}: The aesthetic quality of the music, independent of context.
    \item \textbf{Text-Music Alignment (TMA)}: How effectively the music captures the mood, style, and themes described in the text.
\end{itemize}

Each participant rated a total of 35 music tracks per task (V2M and T2M), completing the questionnaire in approximately 50 minutes. Participants also provide qualitative feedback, explaining their preferences and highlighting emotional or thematic factors influencing their ratings. The reward for each participant is \$10. We finally gather 64 valid responses from social media.

A notable result is observed in the RC metric. Despite not explicitly introducing rhythm-focused features, our model achieves superior performance compared to baseline methods. This suggests that an alternative approach, which emphasizes a broader perspective rather than heavily focusing on local rhythmic features, can also be effective. Overemphasis on local rhythms may sometimes challenge the overall coherence of the music and its alignment with the broader narrative. By adopting a balanced approach, our model maintains rhythmic flow while aligning closely with the intended emotional context, leading to an enhanced audiovisual experience. We also provide a screenshot of the questionnaire with full text of instruction in Fig.~\ref{fig:userstudy}.

\subsection{Controllability Experiment}
\label{sec:control_supp}
To further evaluate the controllability of our model, we test its ability to adjust the tempo of the generated music. Similar to other music attributes such as genre, instrument, and mood, we report the average change in tempo on the generated variations to quantify the model's performance.

For this experiment, we categorize the dataset into distinct beats per minute (BPM) groups: ``Fast,'' ``Medium,'' and ``Slow.'' We randomly sample 20 songs for each BPM group and generate 10 variations for each song, conditioned on the sampled song itself. This setup allows us to assess the model's capability to independently control tempo while maintaining the overall coherence of the generated music.

We calculate the average BPM across the 200 generated songs per group (20 songs $\times$ 10 variations) to measure how well the generated music aligns with the expected tempo adjustments. The results are summarized in Table~\ref{tab:bpm}.

\section{Demos}

Due to PDF limitations, we only showcase the video-to-description generation demo here. Demos for other generation will be uploaded on \url{https://github.com/wbs2788/VMB}.

\label{sec:v2d_samp}
To evaluate our model's performance, we sampled 10 frames from each video and provided the following prompt to the models: \texttt{Based on the emotional tone, pacing, and visuals of this video, how would you compose the music? Provide a one-sentence summary of the key musical elements that you use.} This method ensures that the generated music aligns well with the emotional and visual characteristics of the video. Samples of these music descriptions are displayed in Table~\ref{tab:v2d}. Additionally, the content shown in Figure 1 of the main paper, which illustrates how a poem by P.B. Shelley, ``The Flower that Smiles Today,'' can also be transformed into a music description, is also generated using our model.

\begin{table*}[ht]
\centering
\caption{Samples of visual-to-description generation.}
\label{tab:v2d}
\begin{adjustbox}{max width=0.7\textwidth}
\begin{tabular}{@{}m{6cm}|m{6cm}|m{8cm}@{}}
\toprule
\textbf{Image} & \textbf{Description} & \textbf{GPT-4 Evaluation} \\ \midrule
\adjustbox{valign=c, clip, width=6cm}{\includegraphics[]{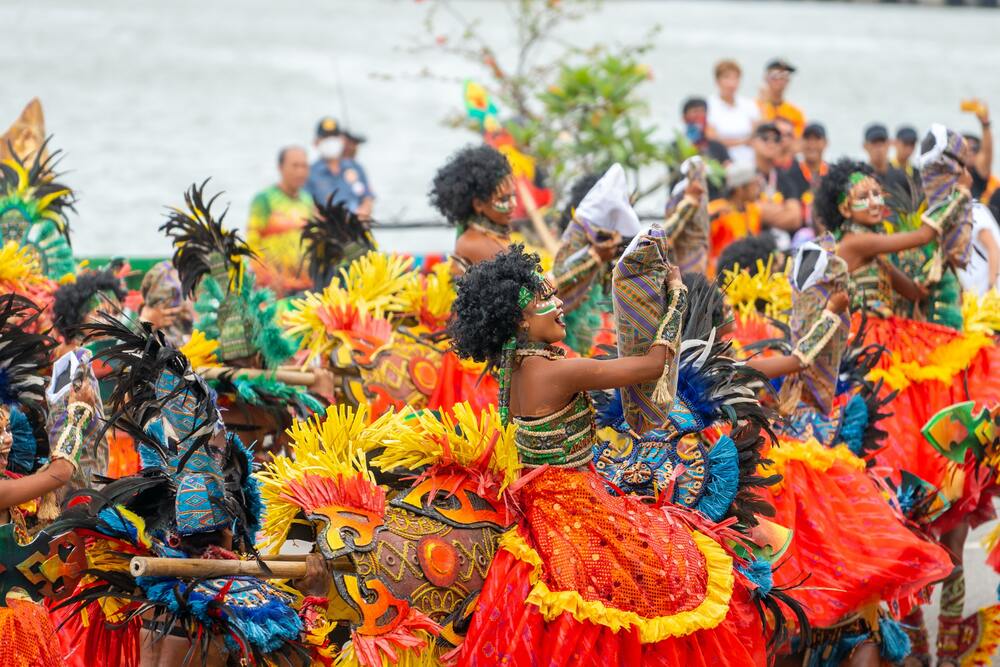}}
& A vibrant, energetic, and epic soundtrack featuring a dynamic blend of strings, brass, and orchestral elements, perfectly capturing a sense of adventure and excitement. & \textbf{Score:} 5 / 5 \newline
\textbf{Reason:} \newline
\textbf{Emotion Match:} The energy and vibrancy of the description match perfectly with the lively festival scene. \newline
\textbf{Scene Association:} The use of ``strings, brass, and orchestral elements'' effectively aligns with the celebratory and grand setting. \newline
\textbf{Conclusion:} The description is highly appropriate for this image, requiring no further improvement.
\\ \hline

\adjustbox{valign=c, clip, width=6cm}{\includegraphics[height=3cm]{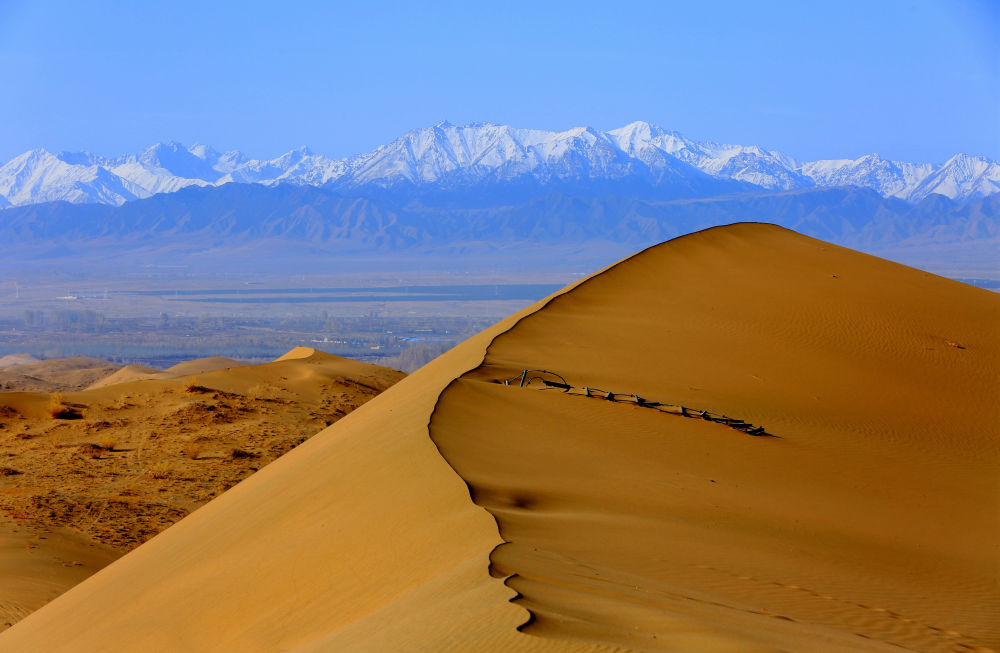} }
& A dynamic and emotive soundtrack featuring a mix of guitar, bass, strings, and orchestra, perfectly capturing a range of moods from action-packed and adventurous to melancholic and romantic. & \textbf{Score:} 4.5 / 5 \newline
\textbf{Reason:} \newline
\textbf{Emotion Match:} The description's adventurous tone works well for the desert landscape. However, the ``melancholic and romantic'' portion slightly detracts from the overall alignment. \newline
\textbf{Scene Association:} The instrumentation (guitar, bass, and orchestra) complements the sense of vastness and exploration in the desert. \newline
\textbf{Conclusion:} A more focused description on adventure and solitude could enhance the fit.\\ \hline

\adjustbox{valign=c, clip, width=6cm}{\includegraphics[height=3cm]{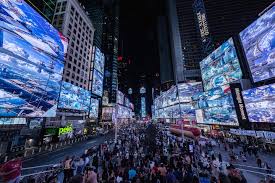}  }
& A fast-paced, dynamic electronic soundtrack featuring pulsating rhythms, vibrant synthwave beats.& \textbf{Score:} 5 / 5 \newline
\textbf{Reason:} \newline
\textbf{Emotion Match:} The fast-paced, electronic, and vibrant tone perfectly reflects the energy, dynamism, and modernity of the urban night scene. \newline
\textbf{Scene Association:} The use of ``synthwave beats'' aligns exceptionally well with the futuristic, neon-lit visuals. \newline
\textbf{Conclusion:} The updated description captures the essence of the image flawlessly, making it an excellent match. \\ \hline

\adjustbox{valign=c, clip, width=6cm}{\includegraphics[height=3cm]{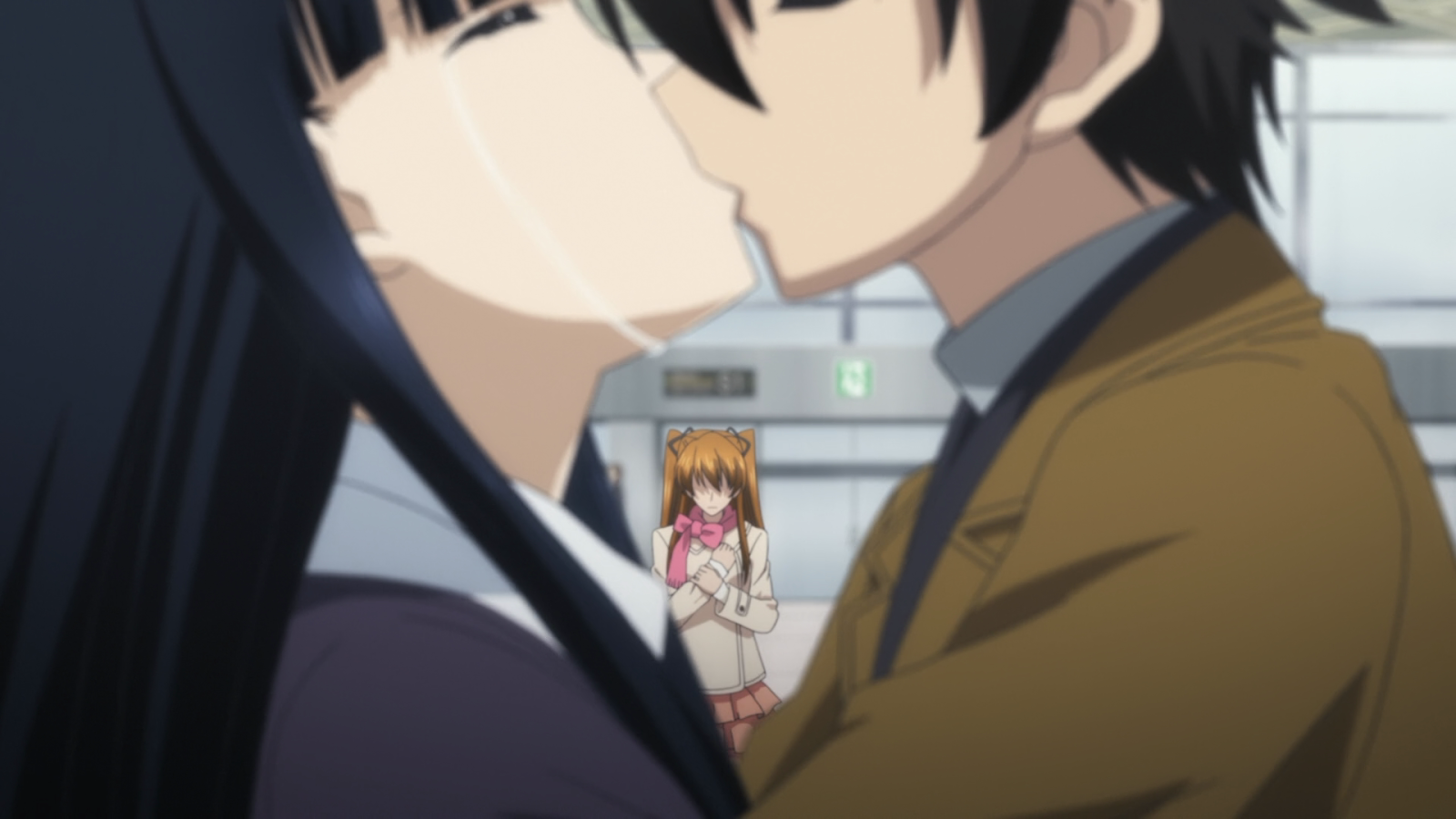} }
&  A gentle piano melody, accompanied by soft strings, to evoke a sense of tenderness.& \textbf{Score:} 5 / 5 \newline
\textbf{Reason:} \newline
\textbf{Emotion Match:} The gentle piano melody perfectly evoke the tenderness and nostalgia expressed in the characters' emotional moment. \newline
\textbf{Scene Association:} The use of ``soft strings'' aligns with the intimate and heartfelt nature of the scene, enhancing the emotional depth. \newline
\textbf{Conclusion:} The description is highly appropriate for this image, requiring no further improvement.\\ \hline

\adjustbox{valign=c, clip, width=6cm}{\includegraphics[height=3cm]{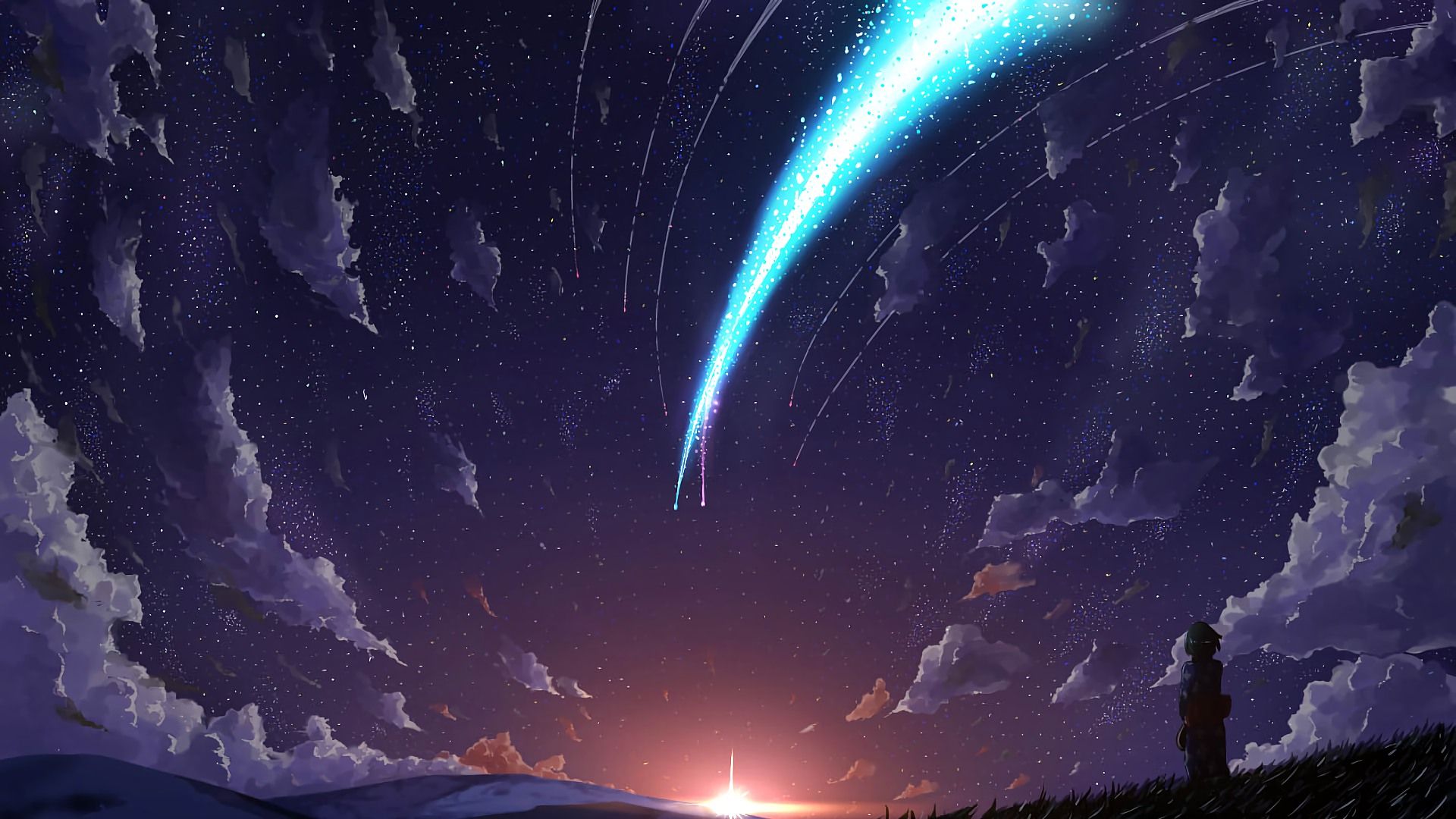}}   &
A gentle, melancholic melody, featuring soft piano and strings, to evoke the serene yet poignant atmosphere.& \textbf{Score:} 4.5 / 5 \newline
\textbf{Reason:} \newline
\textbf{Emotion Match:} The melancholic melody matches the serene and poignant atmosphere of the comet-lit sky. \newline
\textbf{Scene Association:} The inclusion of ``soft piano'' and ``strings'' reflects the calmness and wonder of the scene but does not fully emphasize the awe-inspiring grandeur of the comet. \newline
\textbf{Conclusion:} While the description aligns well with the scene, adding a sense of scale and majesty could enhance the match.
\\ \hline

\adjustbox{valign=c, clip, width=6cm}{\includegraphics[height=3cm]{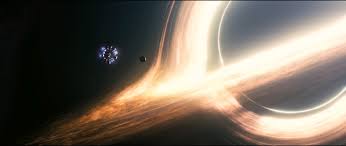}} 
& A sense of awe and wonder, using a combination of orchestral strings, and electronic soundscapes.& \textbf{Score:} 5 / 5 \newline
\textbf{Reason:} \newline
\textbf{Emotion Match:} The energy and vibrancy of the description match perfectly with the lively festival scene. \newline
\textbf{Scene Association:} The use of ``strings, brass, and orchestral elements'' effectively aligns with the celebratory and grand setting. \newline
\textbf{Conclusion:} The description is highly appropriate for this image, requiring no further improvement.
\\ \hline

\adjustbox{valign=c, clip, width=5.5cm}{\includegraphics[height=3cm]{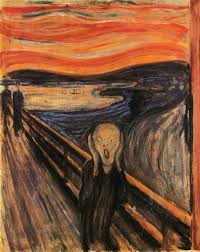}} 
& A slow, eerie, and melancholic melody, using a combination of dissonant chords and a haunting vocal line to evoke the sense of despair and isolation.& \textbf{Score:} 5 / 5 \newline
\textbf{Reason:} \newline
\textbf{Emotion Match:} The slow, eerie melody and dissonant chords align seamlessly with the despair and isolation depicted in \textit{The Scream}. \newline
\textbf{Scene Association:}  The ``dissonant chords'' effectively complements the painting's unsettling and surreal nature. \newline
\textbf{Conclusion:} The description accurately reflects the psychological intensity of the image, requiring no further improvement.
\\

\bottomrule
\end{tabular}
\end{adjustbox}
\end{table*}

\end{document}